\documentclass[twoside,11pt]{article}

\usepackage[accepted]{melba}  
\usepackage[counterclockwise, figuresleft]{rotating}

\usepackage{amsmath,amsfonts}
\usepackage{booktabs}
\usepackage{gensymb}
\usepackage{tabularx}
\usepackage{adjustbox}
\usepackage{hhline}
\usepackage{multirow}

\let\ACMmaketitle=\maketitle
\renewcommand{\maketitle}{\begingroup\let\footnote=\thanks \ACMmaketitle\endgroup}

\melbaheading{015}{https://www.melba-journal.org/article/27657}{2021}{1-40}{01/2021}{09/2021}{Sergiu Mocanu, Alan R. Moody, April Khademi}{}{}

\ShortHeadings{FlowReg: CNN Registration}{Mocanu et al.}
\firstpageno{1}

\title{FlowReg: Fast Deformable Unsupervised Medical Image Registration using Optical Flow\footnote{Data used in preparation of this article were obtained from the Alzheimer's Disease Neuroimaging Initiative (ADNI) database (adni.loni.usc.edu). As such, the investigators within the ADNI contributed to the design and implementation of ADNI and/or provided data but did not participate in analysis or writing of this report. A complete listing of ADNI investigators can be found at: \url{http://adni.loni.usc.edu/wp-content/uploads/how_to_apply/ADNI_Acknowledgement_List.pdf}.}}

\author{\name Sergiu Mocanu \email smocanu@ryerson.ca \\  
	\addr Electrical, Computer, and Biomedical Engineering, Ryerson University, Toronto, ON, Canada
	\AND
	\name Alan R. Moody \email alan.moody@sunnybrook.ca\\
	\addr Department of Medical Imaging, University of Toronto, Toronto, ON, Canada
	\AND
	\name April Khademi \email akhademi@ryerson.ca \\
	\addr Electrical, Computer, and Biomedical Engineering, Ryerson University, Toronto, ON, Canada.\\
	\addr Keenan Research Center for Biomedical Science, St. Michael's Hospital, Unity Health Network, Toronto, ON, Canada.\\
	\addr Institute for Biomedical Engineering, Science and Technology (iBEST), a partnership between St. Michael’s Hospital and Ryerson University, Toronto, ON, Canada.
}

\begin{document}
\maketitle

\begin{abstract}
In this work we propose \textit{FlowReg}, a deep learning-based framework that performs unsupervised image registration for neuroimaging applications. The system is composed of two architectures that are trained sequentially: \textit{FlowReg-A} which affinely corrects for gross differences between moving and fixed volumes in 3D followed by \textit{FlowReg-O} which performs pixel-wise deformations on a slice-by-slice basis for fine tuning in 2D. \textit{FlowReg-A} warps the moving volume using gross global parameters to align rotation, scale, shear, and translation to the fixed volume. A correlation loss that encourages global alignment between the moving and the fixed volumes is employed to regress the affine parameters. The deformable network \textit{FlowReg-O}  operates on 2D image slices and is based on the optical flow CNN network that is adapted to neuroimaging with three loss components. The photometric loss minimizes pixel intensity differences, the smoothness loss encourages similar magnitudes between neighbouring vectors, and a correlation loss that is used  to maintain the intensity similarity between fixed and moving image slices. The proposed method is compared to four open source registration techniques  \textit{ANTs}, \textit{Demons}, \textit{SE}, and \textit{Voxelmorph} for FLAIR MRI applications. In total, $4643$ FLAIR MR imaging volumes (approximately $255,000$ image slices) are used from dementia and vascular disease cohorts, acquired from over 60 international centres with varying acquisition parameters. To quantitatively assess the performance of the registration tools, a battery of novel validation metrics are proposed that focus on the structural integrity of tissues, spatial alignment, and intensity similarity. Experimental results show \textit{FlowReg} (\textit{FlowReg-A+O}) performs better than iterative-based registration algorithms for intensity and spatial alignment metrics with a Pixelwise Agreement (PWA) of $0.65$, correlation coefficient (R) of $0.80$, and Mutual Information (MI) of $0.29$. Among the deep learning frameworks evaluated, \textit{FlowReg-A} or \textit{FlowReg-A+O} provided the highest performance over all but one of the metrics.   Results show that \textit{FlowReg} is able to obtain high intensity and spatial similarity between the moving and the fixed volumes while maintaining the shape and structure of anatomy and pathology.
Our code is available at~\url{https://github.com/IAMLAB-Ryerson/FlowReg}.
\end{abstract}


\section{Introduction}
Magnetic resonance imaging (MRI) offers non-invasive visualization of soft tissue that is ideal for imaging the brain. The etiology and pathogenesis of neurodegeneration, and the effects of treatment options have been heavily investigated in T1, T2, Proton Density (PD), Diffusion Weighted (DW), and Fluid-Attenuated Inversion Recovery (FLAIR) MR sequences  \citep{udaka2002white}\citep{hunt1989clinical}\citep{sarbu2016white}\citep{Tripiii11}\citep{guimiot2008contribution} for dementia \citep{oppedal2015classifying} and Alzheimer's Disease (AD) \citep{kobayashi2002apoptosis}. As cerebrovascular disease (CVD) has been shown to be a leading cause of dementia, there is growing interest into examining cerebrovascular risk factors in the brain using neuroimaging. CVD markers, such as white matter lesions (WML), predict cognitive decline, dementia, stroke, death, and WML progression increases these risks \citep{debette2010clinical}, \citep{alber2019white}. 
Research into the vascular contributions to dementia and neurodegeneration could be valuable  for developing new therapies \citep{frey2019characterization}, \citep{alber2019white}, \citep{griffanti2018classification}, \citep{malloy2007neuroimaging}. FLAIR MRI is preferred for WML analysis \citep{wardlaw2013neuroimaging} \citep{badji2020cerebrovascular} since the usual T2 high signal from cerebrospinal fluid (CSF) is suppressed, highlighting the white matter disease (WMD) high signal. This is due to increased water content secondary to ischemia and demyelination and much more robustly seen in FLAIR than with T1 and T2 sequences \citep{lao2008computer}, \citep{khademi2011robust}, \citep{kim2008classification}, \cite{piguet2005comparing}.\\
%
%
\indent One family of algorithms heavily relied upon in neuroimaging research is image registration, which is the process of aligning two images (one fixed and one moving) so they are in the same geometric space. Structures and changes between two images can be directly compared when images are registered to align longitudinal scans of the same patient to assess disease change over time \citep{csapo2012longitudinal}, \citep{el2016current},  map patient images to an atlas for template-based segmentation \citep{iglesias2015multi}, \citep{phellan2014improving}, or to correct for artifacts such as patient motion and orientation \citep{mani2013survey}. As medical images are composed of highly relevant anatomical and pathological structures such as brain tissue, ventricles, and WMLs, it is important that the shape and relative size of each structure is maintained in the registered output.  MR images are highly complex so obtaining correspondence while maintaining structural and anatomical integrity presents a challenging task.\\
\indent Traditionally, the process of registration, or aligning two images has been framed as an optimization problem, which searches for the transform $T$ between a moving $(I_{m})$ and a fixed $(I_{f})$ image by optimizing some similarity criteria between the fixed and moving images $T = \arg \max _{T} \mathcal{C}\left(I_{f}, T\left(I_{m}\right)\right)$. This optimization can be calculated via gradient descent and ends when maximum similarity is found or a maximum number of iterations is obtained.  The similarity between the fixed and (transformed) moving images is calculated via a cost function $\mathcal{C}$, such as 
mutual information (MI), cross-correlation (CC), and mean-squared error (MSE) \citep{maes1997multimodality} \citep{avants2008symmetric}.
Registrations can be done globally via affine transformation (translation, rotation, shear, and scale) or on a per-pixel level through the use of non-uniform deformation fields (each pixel in the moving image has a target movement vector). \\
\indent
Registration algorithms that involve an iterative-based approach are computationally expensive and any calculated characteristics are not saved after an intensive computational procedure; the transformation parameters are discarded and not used for the next pair of images. In large multi-centre datasets this can create large computation times or non-optimal transformations. To overcome the non-transferable nature of traditional image registration algorithms, machine learning models that learn transformation parameters between images have been gaining interest  \citep{cao2017deformable} \citep{sokooti2017nonrigid} \citep{balakrishnan2018unsupervised}.\\
\indent Recently, several convolutional neural network-based (CNN) medical image registration algorithms have emerged to address the non-transferable nature, lengthy execution and high computation cost of the classic iterative-based approaches \citep{balakrishnan2018unsupervised, uzunova2017training}. In 2017, researchers adapted an optical flow CNN model, \textit{FlowNet} \citep{dosovitskiy2015flownet}, to compute the deformation field between temporally spaced images \citep{uzunova2017training}. Although promising, this approach required ground truth deformation fields during training, which is intractable for large clinical datasets. To overcome these challenges, \textit{Voxelmorph} was developed as a completely unsupervised CNN-based registration scheme that learns a transformation without labeled deformation fields \citep{jaderberg2015spatial}. Further work by Fan et. al \cite{fan2018adversarial} has shown that Generative Adversarial Networks (GAN) can be used to generate deformation fields. These fields are then used to warp the moving image until the discriminator is unable to distinguish between the registered and fixed image. Others have suggested a sequential affine and deformable 3D network for brain MRI registration \cite{zhu2020unsupervised}. In another work, \cite{zhao2019unsupervised} proposed a fully unsupervised method based on CNNs that includes cascaded affine and deformable networks to perform alignment in 3D in one framework. The proposed method is inspired by these pioneering works but instead an affine alignment is performed in 3D first, followed by a 2D fine-tuning on a slice-by-slice basis. \\
\indent Global differences such as head angle or brain size can vary significantly between patients and these global differences are likely to be mainly realized in 3D. Additionally, there are local and fine anatomical differences that are more visible on a per slice basis. Therefore, to get maximal alignment between neuroimages, both global differences in 3D and local differences in 2D should be addressed.  Other design considerations include dependence on ground truth data which is impractical to obtain for large datasets. Lastly, and importantly, registration algorithms must maintain the structural integrity of important objects such as WML and ventricles.\\
%
\indent To this end, this paper proposes a CNN-based registration method called FlowReg: Fast Deformable Unsupervised Image Registration using Optical Flow that addresses these design considerations in a unified framework. \textit{FlowReg} is composed of an affine network \textit{FlowReg-A} for alignment of gross head differences in 3D and a secondary deformable registration network \textit{FlowReg-O} that is based on the optical flow CNNs \citep{ilg2017flownet} \citep{jason2016back} for fine movements of pixels in 2D, thus managing both global and local differences at the same time. In contrast to previous works that perform affine and deformable registration strictly in 3D, we postulate that performing affine registration in 3D followed by 2D refinement will result in higher quality registrations for neuroimages.  \textit{FlowReg} is fully unsupervised, and ground truth deformations are not required.  The loss functions are modified to ensure optimized performance for neuroimaging applications. Lastly, this framework does not require preprocessing such as brain extraction and is applied to the whole head of neuroimaging scans.\\
%
\indent As an additional contribution, a battery of novel validation metrics are proposed to quantify registration performance from a clinically salient perspective. Validation of registration performance is not a simple task and many methods have employed manually placed landmarks \citep{balakrishnan2018unsupervised}, \citep{uzunova2017training}, \citep{de2019deep}. In medical image registration, it is of interest to maintain the structural integrity of anatomy and pathology in the moving images, while obtaining maximum alignment with the fixed volume. To measure this, three groups of metrics are proposed: structural integrity, spatial alignment, and intensity similarity. Structural (tissue) integrity is measured via volumetric and structural analysis with the proportional volume (PV), volume ratios $\Delta V$ and surface-surface distance (SSD) metrics. Spatial alignment is measured via pixelwise agreement (PWA), head angle (HA) and dice similarity coefficient (DSC). Intensity similarity is measured using traditional metrics: mutual information (MI) and correlation coefficient (Pearson's R) as well as one additional new one, the mean-intensity difference (MID).\\
\indent The performance of \textit{FlowReg} is compared to four established and openly available image registration methods: \textit{Demons} \citep{thirion1995fast,thirion1998image,pennec1999understanding}, \textit{Elastix} \citep{klein2010elastix}, \textit{ANTs} \citep{avants2008symmetric}, and \textit{Voxelmorph} \citep{balakrishnan2018unsupervised}. Three are traditional non-learning iterative based registration methods while \textit{Voxelmorph} is a CNN-based registration tool. Performance is measured over a large and diverse multi-institutional dataset collected from over 60 imaging centres world wide of subjects with dementia (ADNI) \citep{mueller2005alzheimer} and vascular disease (CAIN) \citep{tardif2013atherosclerosis}. There are roughly 270,000 images and 4900 imaging volumes in these datasets with a range of imaging acquisition parameters. The rest of the paper is structured as follows: Section \ref{sec:FlowReg} describes the \textit{FlowReg} architecture and Section \ref{sec: validationmetrics} outlines the validation metrics. The data used, experimental setup, and results are shown in Section \ref{sec: experimentsandresults} followed by the discussions and conclusions in Section \ref{sec:discussion} and Section \ref{sec:conclusion}.
\section{Methods}
In this section, we introduce \textit{FlowReg}, Fast Deformable Unsupervised Image Registration using Optical Flow, with focus on neuroimaging applications. Given a moving image volume denoted by $M(x,y,z)$, where $M$ is the intensity at the $(x,y,z)\in Z^3$ voxel, and the fixed image volume $F(x,y,z)$ the system automatically learns the registration parameters for many image pairs, and uses that knowledge to predict the transformation $T$ for new testing images. The section closes with  a battery of novel registration validation metrics focused on structural integrity, intensity and spatial alignment.
\begin{figure}
    \centering
    \includegraphics[width=.75\textwidth]{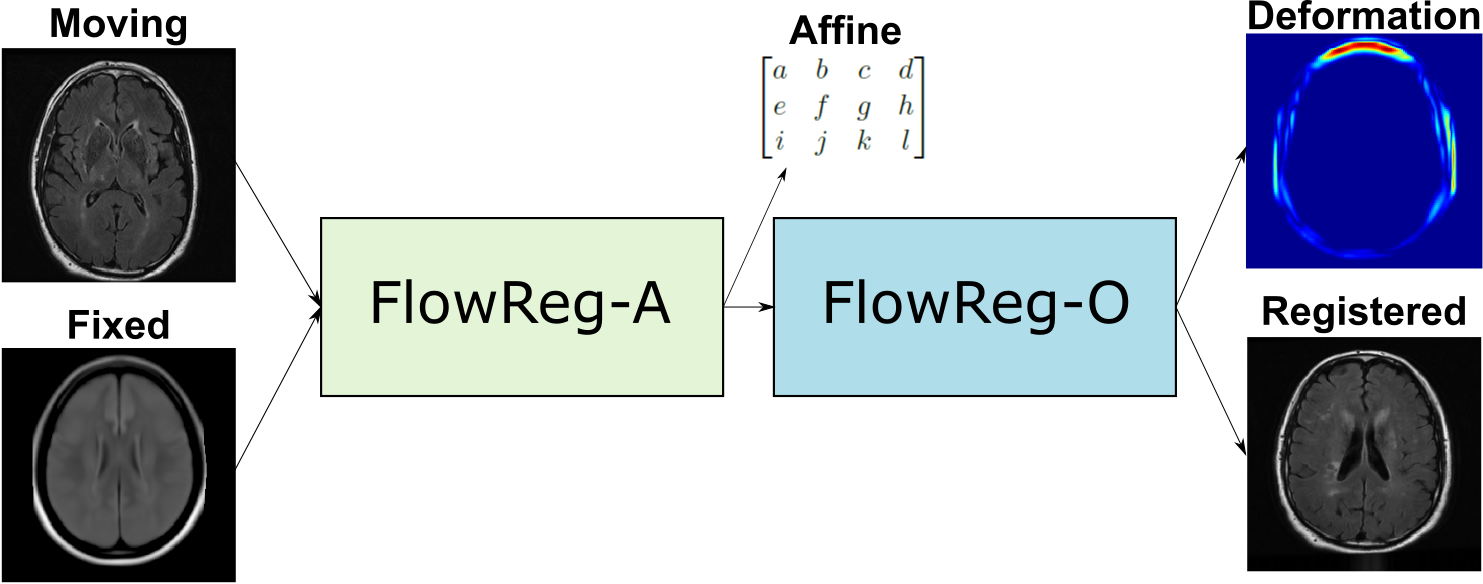}
    \caption{\textit{FlowReg} consists of affine (\textit{FlowReg-A}) and optical flow (\textit{FlowReg-O}) networks.}
    \label{fig:FlowRegWhole}
\end{figure}
\subsection{FlowReg}\label{sec:FlowReg}
\textit{FlowReg}, is based exclusively on CNN architectures, and the alignment is fully unsupervised, meaning that registration parameters are regressed without ground truth knowledge. Registration with \textit{FlowReg} is completed with a two phase approach shown in Fig. \ref{fig:FlowRegWhole}.  \textit{FlowReg-A} is a  3D affine registration network that corrects for global differences and \textit{FlowReg-O} refines the affine registration results on a per slice basis through  a 2D optical flow-based registration method from the video processing field \citep{dosovitskiy2015flownet}, \citep{ilg2017flownet}, \citep{garg2016unsupervised}, \citep{jason2016back}. The affine component is trained first and the affine parameters are obtained for each volume. Once all the volumes are registered using the affine components, \textit{FlowReg-O} is trained to obtain the deformation fields. The held out test set is used to test the full FlowReg pipeline end-to-end.
\subsubsection{FlowReg-A: Affine Registration in 3D}
\begin{figure}
    \centering
    \includegraphics[width=0.8\textwidth]{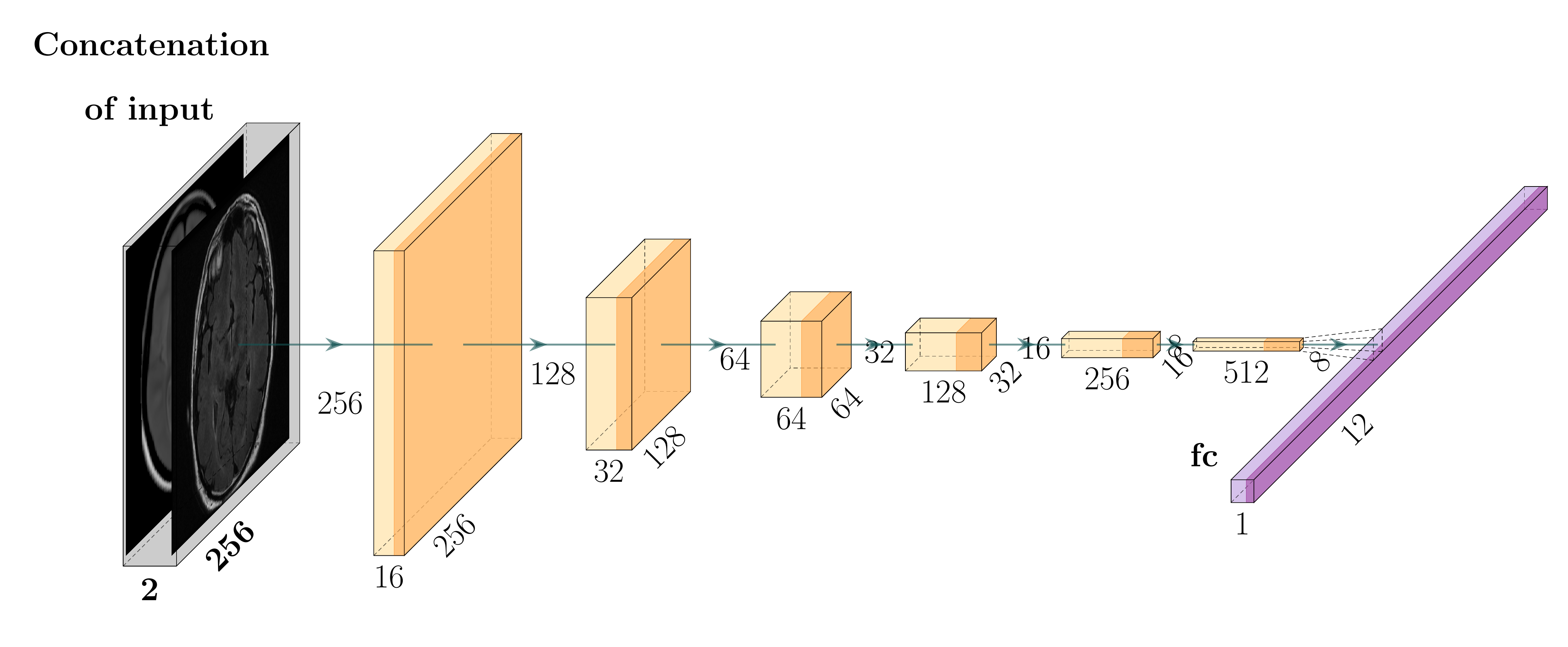}
    \caption{\textit{Flowreg-A} model structure. A pair of 3D input volumes are concatenated, each yellow box represents the output of 3D convolutional layers, the numbers at the bottom of the box are the number of feature maps generated by each convolutional kernel. The last layer (purple) is a fully connected layer with 12 nodes and a linear activation function.}
    \label{fig:FlowReg-A}
\end{figure}
\indent The proposed affine model \textit{FlowReg-A} warps the moving volume using gross global parameters to align head rotation, scale, shear, and translation to the fixed volume. This is beneficial when imaging the brain in 3D, since each the orientation of subjects' heads can vary. Additionally, images are often of different physical dimensions depending on the scanner type and parameters used for acquisition. To normalize these global differences, we propose a completely unsupervised CNN-based 3D affine registration method  (i.e.\ volume registration), where the transformation parameters are learned.\\
\indent The CNN network used to regress the affine matrix parameters is shown in Figure \ref{fig:FlowReg-A} and described in Table \ref{tab:flowreg-a}. The network architecture and hyperparameter selection is similar to the encoder arm of the \textit{FlowNet-Simple} network, with changes made to the input size and the number of 3D convolution kernels. The affine model is comprised of six convolutional layers and one fully-connected layer which is used to regress the flattened version of the three-dimensional affine matrix, $A$:
\begin{equation}
A=
    \begin{bmatrix}
    a & b & c & d\\
    e & f & g & h\\
    i & j & k & l
    \end{bmatrix},
\end{equation}
where $A$ contains the rotation, scale, and translation parameters. Given this affine transformation matrix, the original image volume may be transformed by $M_w(x,y,z) = A \times M(x,y,z)$. \\
\indent To solve for the transformation parameters $A$, a correlation loss was used to encourage an overall alignment of the mean intensities between the moving and the fixed volumes:
\begin{equation}
\label{eqn:correlation_loss_3d}
\begin{split}
    \ell_{\operatorname{corr}_{3 D}}\left(F, M_{w}\right)=1-\frac{\sum_{i=1}^{N}\left(F_{i}-\overline{F}\right)\left(M_{w_{i}}-\overline{M_{w}}\right)}{\sqrt{\sum_{i=1}^{N}\left(F_{i}-\overline{F}\right)^{2}} \sqrt{\sum_{i=1}^{N}\left(M_{w_{i}}-\overline{M_{w}}\right)^{2}}}
\end{split}
\end{equation}
where $F$ is the fixed volume, $M_w$ is the moving volume warped with the calculated matrix, $N$ is the number of voxels in the volume, $F_i$ is the $i^{th}$ element in $F$, $M_{w_{i}}$ is the $i^{th}$ element in $M_w$, and $\overline{F}$, $\overline{M_w}$ are the mean intensities of the fixed and moving volumes, respectively. Using the correlation loss function, the parameters are selected during training that ensures global alignment between the fixed and moving image volumes.
\subsubsection{FlowReg-O: Optical Flow-based Registration in 2D}
\begin{figure}
    \centering
    \includegraphics[width=\textwidth]{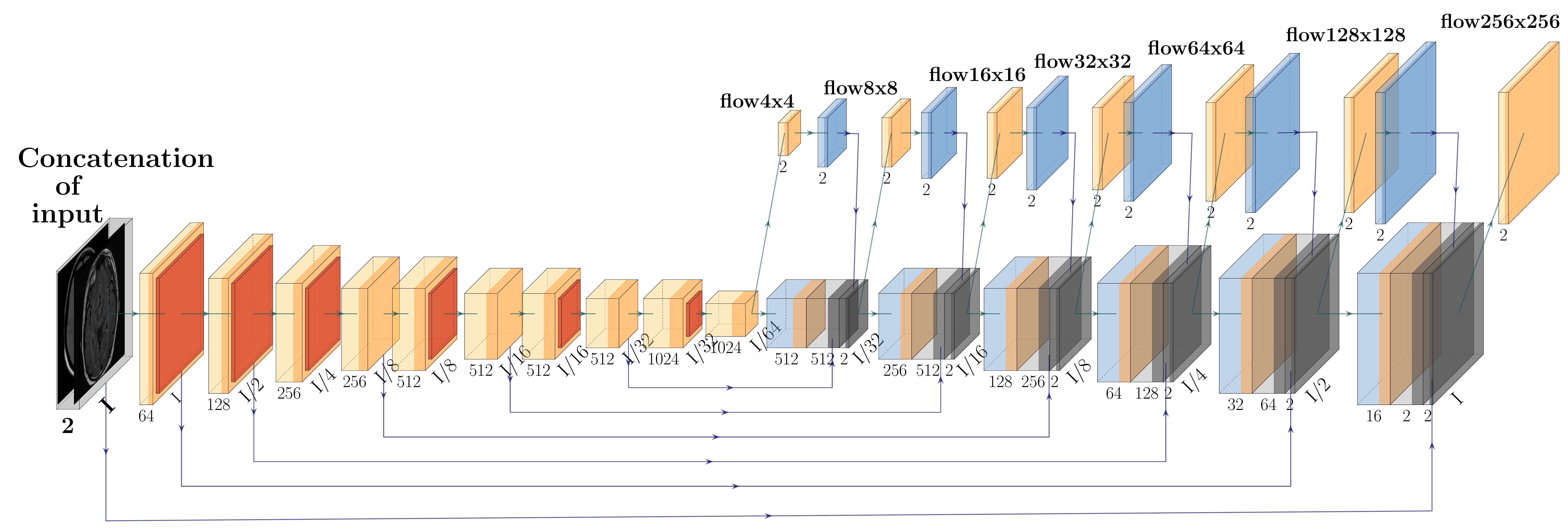}
    \caption{\textit{Flowreg-O} model structure. A pair of 2D input images are concatenated first followed by 2D convolutional layers (yellow). Numbers below layers correspond to the number of feature maps. Skip connections between the upscaling decoder (blue) arm are concatenated (gray boxes) with the output of the encoder layers. The flow at seven resolutions are labeled with \textit{flow} above the corresponding outputs. \textit{I} is the input image resolution $256\times256$.}
    \label{fig:FlowReg-O}
\end{figure}

\begin{figure}
    \centering
    \includegraphics[width=0.5\textwidth]{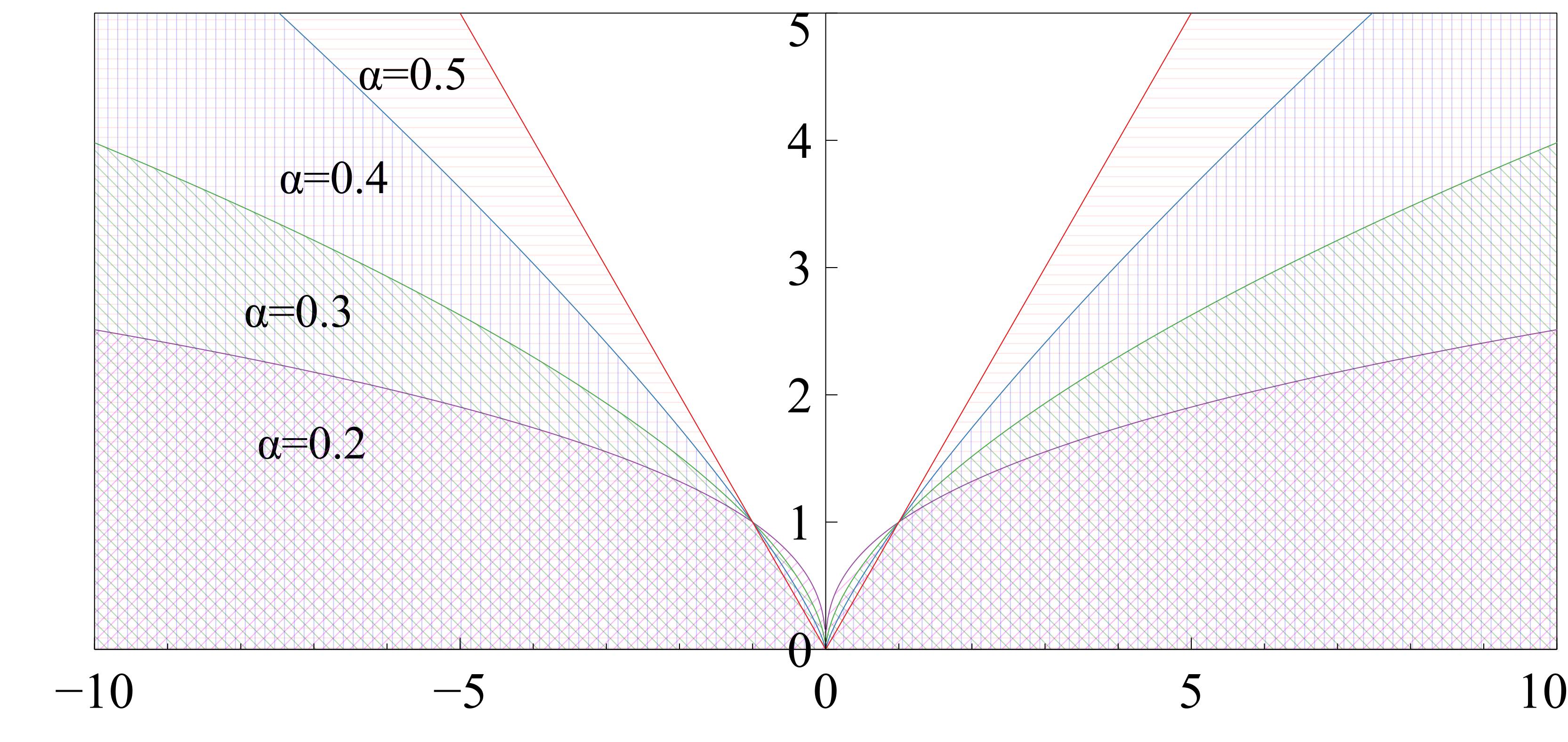}
    \caption{Charbonnier function (Eqn. \ref{eqn: charbonnier}) for $\alpha =$ $0.5$, $0.4$, $0.3$, and $0.2$.}
    \label{fig:charbonnier}
\end{figure}

\begin{figure}[htp!]
    \centering
    \includegraphics[width=0.75\textwidth]{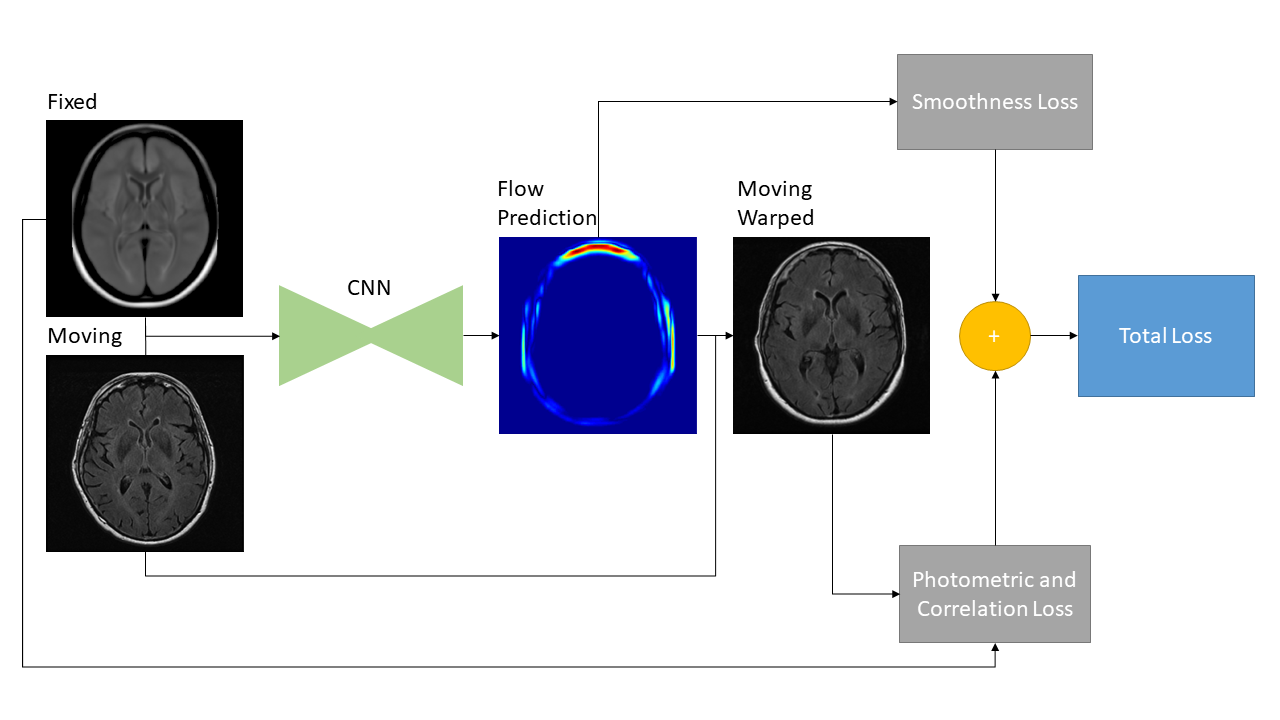}
    \caption{Overview of loss components for the deformable registration network, FlowReg-O.}
    \label{fig:loss components}
\end{figure}

The optical flow component of the registration framework, \textit{FlowReg-O}, is used to perform fine-tuning of the affine registration results on a slice-by-slice basis (i.e. in 2D). \textit{FlowReg-O} is an adapted version of the original optical flow \textit{FlowNet} architecture, used in video processing frameworks. Optical flow is a vector field that quantifies the apparent displacement of a pixel between two temporally separated images. A video is composed of a number of frames at a certain frame-rate per second, $\frac{F}{s}$ and the optical flow measures the motion between objects and pixels across frames and can be used to calculate the velocity of objects in a scene \citep{horn1981determining}. For the task of medical image registration, instead of aligning neighbouring frames, we will be aligning moving and fixed images. The same principles as the original optical flow framework are adopted here, where the  displacement vector is found and used to warp  pixels between moving and fixed images in 2D.\\
%
\indent The proposed deformable registration network is identical to the \textit{FlowNet Simple} architecture in terms of the convolutional layers and hyperparameter selection, but adapted to grayscale medical image sizes and content. See Fig \ref{fig:FlowReg-O} for the FlowReg-O architecture, which is also described in Table \ref{tab:flowreg-o}.  The original \textit{FlowNet} architecture was implemented on the synthetically generated dataset \textit{"Flying Chairs"} with known optical flow values for ground truths, thus dissimilarities are calculated as a simple endpoint-error (EPE) \citep{dosovitskiy2015flownet,ilg2017flownet}. Since then, unsupervised methods have been proposed to train optical flow regressor networks based on a loss that compares a warped image using the regressed flow and its corresponding target image with the use of Spatial Transformer Networks \citep{jaderberg2015spatial} \citep{jason2016back}.  In medical imaging, the optical flow ground truths are impossible to obtain, so the same unsupervised principle is adopted here, but the losses have been conformed for medical images.  In addition to photometric and smoothness loss components which were used in the original work \citep{jason2016back}, \textit{FlowReg-O} utilizes an additional correlation loss term, with each loss encouraging overall similarity between the fixed and moving images while maintaining small 2D movements in the displacement field.\\
%
\indent The total loss function is a summation of three components: photometric loss $\ell_{photo}$ to keep photometric similarity through the Charbonnier function, the smoothness loss $\ell_{smooth}$ which ensures the deformation field is smooth (and limits sharp discontinuities in the vector field), and the correlation loss $\ell_{corr}$, which was added to enforce global similarity in the intensities between the moving and fixed images. The total loss for FlowReg-O is
\begin{equation}    
\label{eqn: total loss}
\begin{split}
    \mathcal{L}(\mathbf{u,v} ; F(x, y), M_w(x, y)) = &\gamma \cdot \ell_{photo}(\mathbf{u,v}; F(x, y), M_w(x, y)) + \\
    &\zeta \cdot  \ell_{corr}(\mathbf{u,v}; F(x, y), M_w(x, y)) + \lambda \cdot  \ell_{smooth}(\mathbf{u,v})
\end{split}
\end{equation}
where $\mathbf{u,v}$ are the estimated horizontal and vertical vector fields, $F(x,y)$ is the fixed image, $M_w(x,y) = M(x+u,y+v)$ is the warped moving image, and $\gamma$, $\zeta$, and $\lambda$ are weighting hyper-parameters.\\
\indent The photometric loss, adopted from \cite{jason2016back}, is the difference between intensities of the fixed image and the warped moving image and evaluates to what degree the predicted optical flow is able to warp the moving image to match the intensities of the fixed image on a pixel-by-pixel basis:
\begin{equation}
\label{eqn: photometric loss}
\begin{split}
    \ell_{photo}(\mathbf{u,v};&F(x,y),M_w(x,y)) = \frac{1}{N}\sum_{i,j}\rho(F(i,j)-M_w(i,j)))
\end{split}
\end{equation}
where $N$ is the number of pixels and $\rho$ is the Charbonnier penalty function which is used to reduce contributions of outliers. The Charbonnier penalty is defined by:
\begin{equation}
\label{eqn: charbonnier}
    \rho(x) = (x^2 + \epsilon^2)^\alpha
\end{equation}
where $x = (F-M_w)$, $\epsilon$ is a small value ($0.001$), and $\alpha$ regulates the difference in intensities between the moving and fixed images such that large differences can be damped to keep the magnitudes of the deformation vectors within a reasonable limit. The effect of the $\alpha$ parameter on the Charbonnier function is shown in Fig. \ref{fig:charbonnier}.  For smaller $\alpha$ values, the Charbonnier function suppresses the output magnitude which is used to regress finer movements in the displacement field.\\ 
\indent The smoothness loss is implemented to regularize the flow field. The loss component encourages small differences between neighbouring flow vectors in the height and width directions and is defined by
\begin{equation}
\begin{split}
\label{eqn: smoothness loss}
    \ell_{smooth}(\mathbf{u,v}) = \sum_{j}^{H}\sum_{i}^{W}[&\rho(u_{i,j}-u_{i+1,j}) + \rho(u_{i,j} - u_{i,j+1}) + \\
    &\rho(v_{i,j} - v_{i+1,j}) + \rho(v_{i,j} - v_{i,j+1})],
\end{split}
\end{equation}
where $H$ and $W$ are the number of rows and columns in the image and $u_{i,j}$ and $v_{i,j}$ are displacement vectors for pixel $(i,j)$ and $\rho$ is the Charbonnier function.  This loss measures the difference between local displacement vectors and minimizes the chances of optimizing to a large displacement between neighbouring pixels.\\
\indent Lastly, we added an additional correlation loss component to encourage an overall alignment of the mean intensities between the moving and the fixed 2D images (similar to \textit{FlowReg-A}), as in:
\begin{equation}
\label{eqn:correlation_loss_2d}
\begin{split}
    \ell_{\operatorname{corr}_{2 D}}\left(F, M_{w}\right)=1-\frac{\sum_{i=1}^{N}\left(F_{i}-\overline{F}\right)\left(M_{w_{i}}-\overline{M_{w}}\right)}{\sqrt{\sum_{i=1}^{N}\left(F_{i}-\overline{F}\right)^{2}} \sqrt{\sum_{i=1}^{N}\left(M_{w_{i}}-\overline{M_{w}}\right)^{2}}},
\end{split}
\end{equation}
where $F$ is the fixed image, $M_w$ is the moving image warped with the calculated flow, $N$ is the number of pixels in the image, $F_i$ is the $i^{th}$ element in $F$, $M_{w_{i}}$ is the $i^{th}$ element in $M_w$, and $\overline{F}$, $\overline{M_w}$ are the mean intensities of the fixed and moving images, respectively. A summary of the loss components and how they are  implemented for the FlowReg-O network is is shown in Fig. \ref{fig:loss components}. \\
\subsection{Validation Metrics} \label{sec: validationmetrics}
There are three categories of metrics that are proposed that each measure a particular aspect of registration accuracy that have clinical relevance, including: structural integrity, spatial alignment, and intensity similarity. The validation measures for each category are shown in Table \ref{tab:validat_metrics_equations)} and the flow diagram to compute each of the metrics is shown in Figure \ref{fig:process}. In total, there are nine metrics computed, and each metric is summarized in Table \ref{tab:validat_metrics_equations)}. A more detailed explanation of how to compute each metric is available in Appendix \ref{sec:appendixB}.   
%
%

\begin{table}[!h]
    \centering
    \caption{Summary of validation metrics. $M(x,y,z)$, $F(x,y,z)$ and $A(x,y,z)$ are the moving, fixed, and generated atlas volumes respectively, with spatial coordinates $(x,y,z)$. $vol_s$ is the volume of a structure $s$, $N$ is the number of images, $b_M$ and $b_F$ are the brain masks of the moving and fixed, and $p$, $q$ and $r$ are distances between pixels.}
    \begin{tabular}{lll}
    \toprule
     & Metric       & Equation            \\
    \midrule
    \multirow{3}{*}{\rotatebox{-90}{Structural Integrity}} & Proportional Volume & $PV = \frac{vol_s}{vol_b}$ \\
    \\
    & Volume Ratio & $\Delta V_s = \frac{vol_{orig}}{vol_{reg}}$ \\
    \\
    & Surface-Surface-Distance & SSD = $\frac{1}{N}\left[\sum_{i=1}^{N}\mathrm{argmin}_{(x_s,y_s,z_s)\in S} \left(\sqrt{p + q + r}\right) \right]$ \\
    \\ \midrule
    \multirow{3}{*}{\rotatebox{-90}{Spatial Alignment}}& Head Angle & $\theta (\degree$) from midsagital plane\\
    \\
    & Pixel-Wise Agreement & $PWA(z) = \frac{1}{N_j}\frac{1}{N_{xy}} \sum_{j\in J} \sum_{(x,y)}(M_j(x,y,z)-F(x,y,z))^2$ \\
    \\
    & Brain-DSC & $D S C=\frac{2|b_{M} \cap b_{F}|}{|b_{M}|+|b_{F}|}$ \\
    \\ \midrule
    \multirow{3}{*}{\rotatebox{-90}{Intensity Similarity}} & Mutual Information & $I(M;F) = \sum_{f\in F}\sum_{m\in M} p_{(M,F)} (m,f) \log \left( \frac{p_{(M,F)}(m,f)}{p_{M}(m)  p_{F}(f)} \right)$ \\
    \\
    & Correlation Coefficient & $r(M,F) = \frac {\sum_{i=1}^{n} (M_i - \overline{M})(F_i -\overline{F})}{\sqrt{\sum_{i=1}^{n}(M_i - \overline{M})^2} \sqrt{\sum_{i=1}^{n}(F_i - \overline{F})^2}}$ \\
    \\
    & Mean Absolute Intensity Difference & $MAID(A,F) = \frac{1}{N_i}\sum_i|p_A(i)-p_{F}(i)|$ \\
\\ \bottomrule
\end{tabular}
\label{tab:validat_metrics_equations)}
\end{table}
\section{Experiments and Results} \label{sec: experimentsandresults}
In this section the data and the experimental results are detailed.
\subsection{Data}
The performance of FlowReg is evaluated in a large and  diverse FLAIR MRI data repository. Over 270,000 FLAIR MR images were retrieved from two datasets which comprises roughly 5000 imaging volumes from over 60 international imaging centres.  This comprises one of the largest FLAIR MRI datasets in the literature that is being processed automatically to the best of our knowledge.  The first dataset is from the  Canadian Atherosclerosis Imaging Network (CAIN) \cite{tardif2013atherosclerosis} and is a pan-Canadian study of vascular disease. The second dataset is from the Alzheimer's Disease Neuroimaging Initiative (ADNI) \citep{mueller2005alzheimer} which is an international study for Alzheimer's and related dementia pathologies. The acquisition and demographics information is shown in the Appendix in Table \ref{tab:scanners} and \ref{tab:demographics}. Based on image quality metrics supplied  with the ADNI database, scans with large distortions, motion artifacts, or missing slices, were excluded from the study. In total there were 310 volumes excluded based on this criteria. For training, validation and testing an 80/10/10 data split was employed and volumes were randomly sampled from CAIN and ADNI.  The resulting splits were 3714 training volumes (204,270 images), 465 validation volumes (25,575 images) and 464 test volumes (25,520 images).  See Figure \ref{fig:slices_alpha} for example slices from several volumes of the ADNI and CAIN test set, exhibiting wide variability in intensity, contrast, anatomy and pathology.\\
\indent To measure the validation metrics proposed in Section \ref{sec: validationmetrics}, two sets of images are required. Firstly, all volumes in the test set (464 volumes) are used to compute the intensity and spatial alignment metrics: HA, PWA, DSC, MI, Corr, MAID. Second, to compute the structural integrity metrics (PV, volume ratio and SSD metrics), binary segmentation masks of the structures of interest are required and 50 CAIN and 20 ADNI volumes were sampled randomly from the test set for this task. For the objects of interest,  ventricles and WML objects are selected since they represent clinically relevant structures that characterize neurodegeneration and aging. Manual segmentations for the ventricular and WML regions were generated by a medical student trained by a radiologist. These objects are used to examine the structural integrity before and after registration. To generate brain tissue masks, the automated brain extraction method from \citep{khademi2020whole} is utilized to segment cerebral tissue in FLAIR MRI. The atlas used is this work as the fixed volume $F(x,y,z)$ has the dimensions of $256\times256\times55$ and is further detailed in \citep{winklerflair}. The moving image volumes $M(x,y,z)$ comes from the image datasets described in Table \ref{tab:scanners} and no pre-processing was done to any of the volumes other than resizing $M(x,y,z)$ to the atlas resolution ($256\times256\times55$) through bilinear interpolation.
\subsection{Experimental Setup}
\textit{FlowReg-A} and \textit{FlowReg-O} models were trained sequentially. First,  \textit{FlowReg-A} is trained using 3D volume pairs of $M(x,y,z)$ and $F(x,y,z)$ and using the optimized model parameters, the training volume set is affinely registered to the atlas using the found transformation \citep{dalca2018unsupervised}. Subsequently, the globally aligned volumes are used to train \textit{FlowReg-O}, on a slice-by-slice basis, using paired moving $M(x,y)$ and fixed $F(x,y)$ images to obtain the fine-tuning deformation fields in 2D. For \textit{FlowReg-A} and \textit{FlowReg-O} the Adam optimizer was used \citep{kingma2014adam} with a $\beta_1=0.9$ and $\beta_2=0.999$, and a learning rate of $lr=10^{-4}$. \textit{FlowReg-A} training was computed for 100 epochs using a batch size of four pairs of volumes from $M(x,y,z)$ and $F(x,y,z)$. \textit{FlowReg-O} was trained using the globally aligned 2D images for 10 epochs using a batch size of 64 image pairs (2D) at seven two-dimensional resolutions: $256\times256$, $128\times128$, $64\times64$, $32\times32$, $16\times16$, $8\times8$, and $4\times4$. The loss hyper-parameters were set as $\gamma=1$, $\zeta=1$, and $\lambda=0.5$ as per the original optical flow work \cite{jason2016back}. During the testing phase, the deformation field in the last layer of the decoding arm is used to warp the moving test images as this resolution provides per pixel movements and is generated at the same resolution of the input image. Using the trained models for the complete \textit{FlowReg} pipeline, the testing performance is compared to that of \textit{VoxelMorph}, \textit{ANTs}, \textit{Demons} and \textit{SimpleElastix}.\\
\indent Training for CNN models was performed using a NVIDIA GTX 1080Ti, with Keras \citep{chollet2015keras} as a backend to Tensorflow \citep{tensorflow2015-whitepaper} for \textit{FlowReg-A}, \textit{FlowReg-O}, and \textit{Voxelmorph}. \textit{ANTs} registration was performed in Python using the Symmetric Normalization (SyN) with default values \citep{avants2008symmetric}. \textit{Demons} algorithm was implemented in Python using SimpleITK \citep{insight}. Similarly, the Pythonic implementation of Elastix \citep{klein2010elastix} was employed for  \textit{SimpleElastix} \citep{marstal2016simpleelastix} as an add-on to SimpleITK. As a preprocessing step, prior to running \textit{Voxelmorph}, the volumes were first affinely registered using \textit{ANTs-Affine}.  \textit{Voxelmorph} was then trained for 37,000 iterations (to avoid observed overfitting) using the same training dataset utilized for training \textit{FlowReg}. \textit{Voxelmorph} performs 3D convolutions, thus resizing was necessary to keep the output of the decoder the same size as the pooling and upsampling layers. Both the atlas and the moving volumes were resized to  $256\times256\times64$. This resolution was chosen to be a binary multiple of $2^{n}$ to ensure that the decoder arm of the \textit{U-net style} network is able to rebuild the learned feature maps to the original input size and warp the volumes accordingly.\\
\indent To validate the registration methods, the metrics described in Section \ref{sec: validationmetrics} and shown in Fig. \ref{fig:process} were used. The structural integrity validation metrics (PV, volume ratio and SSD) used binary masks for the corresponding brain, ventricle, and WML masks and the resultant deformation field or transformation from each registration method. The PV calculation includes PV from the ventricles and WMLs with respect to the whole brain volume.  The SSD is computed between the ventricle surface and the brain surface only; WML are not included for this metric since small lesion loads and irregular WML boundaries can create large differences in the SSD which may not be related to overall integrity. Finally, for all registration methods and all test data, the large-scale metrics are computed: HA, PWA, DSC, MI, R and MAID were calculated between the registered volumes and atlas for all testing volumes. Warped masks were binarized with a threshold of 0.1 so as to avoid the non-integer values obtained from interpolation.
\subsection{Results}
\begin{figure}
    \centering
    \includegraphics[width=0.5\textwidth]{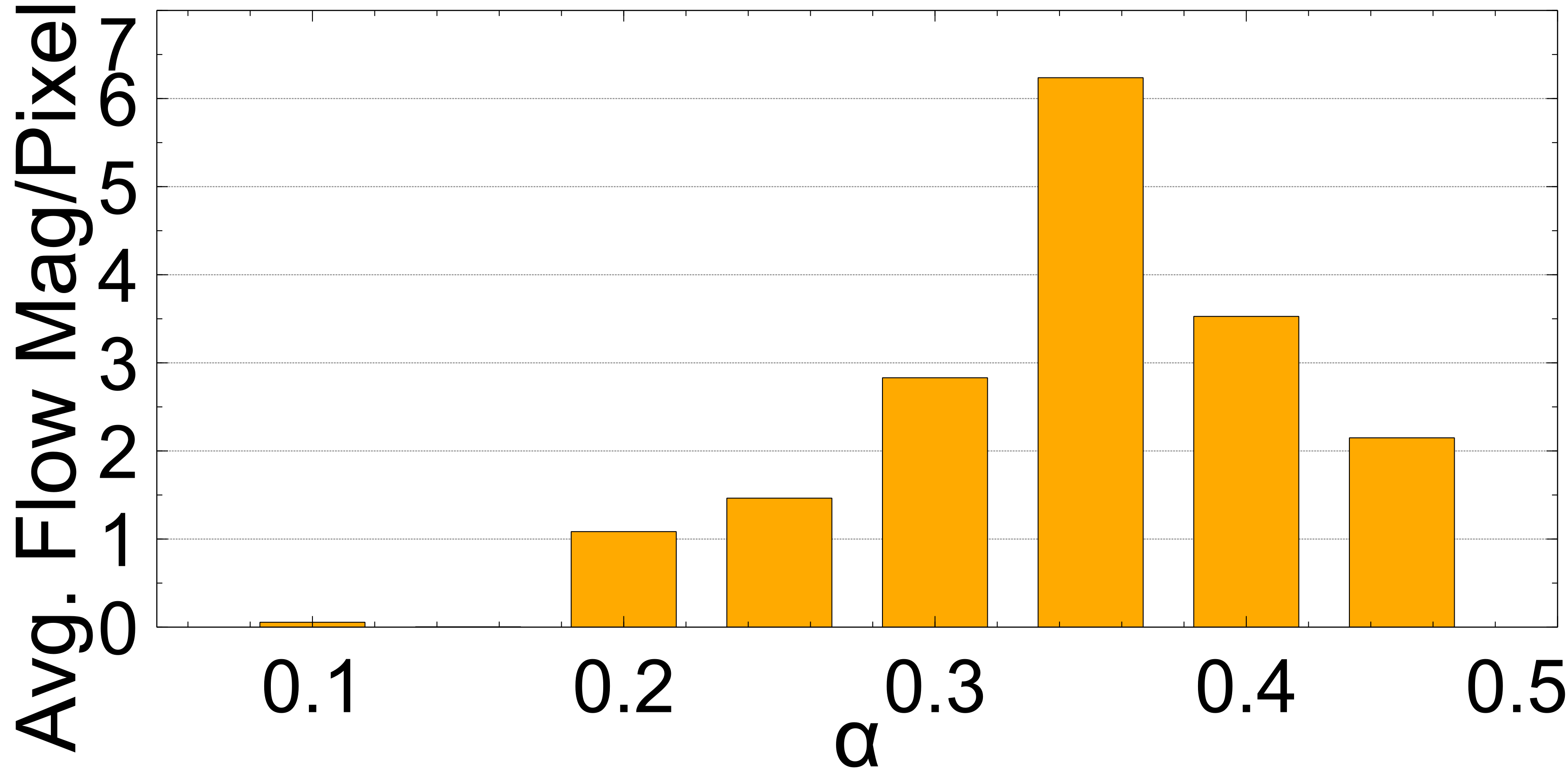}
    \caption{The average flow magnitude per pixel for \textit{FlowReg-O} at various $\alpha$ values.}
    \label{fig:flow_mag_val}
\end{figure}

In this section the experimental results will be presented. First, the effect of $\alpha$ on the Charbonnier penalty function (Equation \ref{eqn: charbonnier}) for the optical flow photometric and smooth loss functions in  \textit{FlowReg-O} was analyzed since this parameter plays a major role in reducing over-fitting and overtly large deformations. Using the held out validation set, the value for $\alpha$ is selected based on the effect this parameter has on the images, which includes visual analysis of the distortion on the images, the magnitude of the optical flow field as well as the average flow magnitude. The model with the appropriate $\alpha$ is used to train the final model for the remainder experiments.\\
%
\indent The registration performance of \textit{FlowReg-O} is studied for $\alpha$ from $0.10$ to $0.45$ in $0.05$ increments by training the network, registering images from the holdout set, and visual inspection. See Figure \ref{fig:slices_alpha} and Figure \ref{fig:flow_mag_img} for images and flow magnitudes for different $\alpha$ in Appendix A. At $\alpha\geq0.25$ values there is more distortion and smearing within the brain, ventricle shapes are warped, and there is distortion of the WMLs. At $\alpha\leq0.15$ there was little to no pixel movement.  These findings are confirmed by computing the average flow magnitude per pixel in Figure \ref{fig:flow_mag_val}, which shows low pixel movement for low $\alpha$ and large displacements for larger $\alpha$. Based on these findings, the "sweet-spot" for $\alpha$ value lies within $0.2$ and $0.25$ to ensure moderate pixel movement without distortion. To ensure there are no overt movements and to be conservative side, we have selected $\alpha = 0.2$ for \textit{FlowReg-O}. The overall effect of \textit{FlowReg-A} and \textit{FlowReg-O} with $\alpha = 0.2$ are used for the final model.
\begin{figure}
    \centering
    \includegraphics[width=\textwidth]{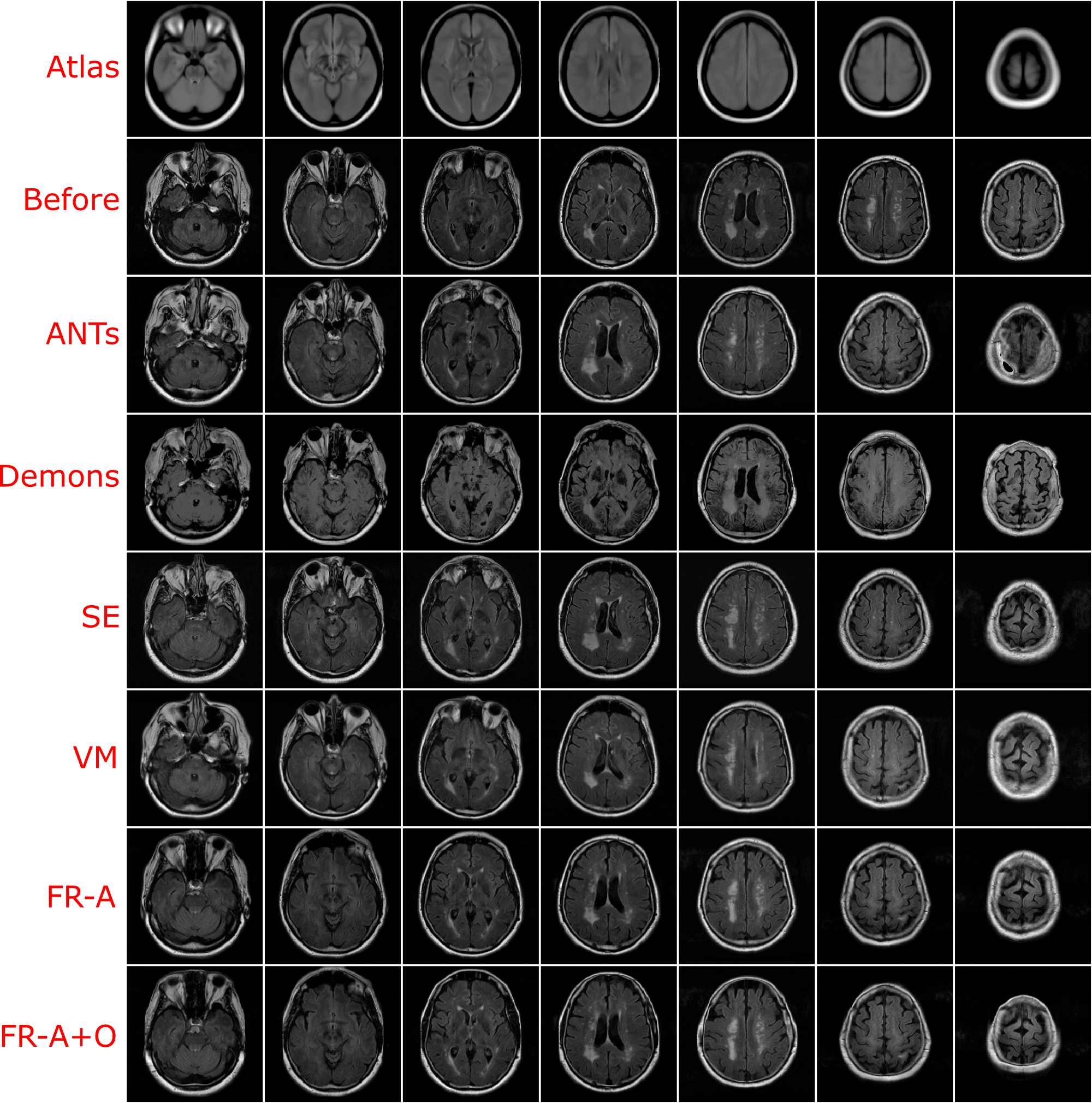}
    \caption{Registration results for seven slices from a single volume. Top row is the atlas that is used as the fixed volume $F(x,y,z)$ and the second row contains the moving volume $M(x,y,z)$. The remaining rows show registration results of ANTs, Demons, SimpleElastix, VoxelMorph, FlowReg-A and FlowReg (FlowReg-A+O).}
    \label{fig:vol_montage}
\end{figure}
\indent Using the finalized \textit{FlowReg} model, test performance was compared to of each of the other registration methods using the proposed validation metrics. All of the testing data (464 volumes) were registered using the models and algorithms described previously.  Example registration results for all of the methods are shown in Figure \ref{fig:vol_montage}.  Bottom, top and middle slices were chosen to show the spectrum of slices that need to be accurately warped and transformed from a 3D imaging perspective.  In the first row,  slices from the fixed (atlas) volume $F(x,y,z)$ are shown, followed by the corresponding slices from the moving volume $M(x,y,z)$. The first column contains images with the ocular orbits and cerebellum in both in the moving and fixed. In the middle slices of the volume, the ventricles are visible and some periventricular WMLs as well. The top slice of the fixed volume is the top of the head and is included since it is comprised of small amounts of brain tissue. The remaining rows display the results of registering the moving images $M(x,y,z)$ to the fixed images $F(x,y,z)$ for each of the respective tools.\\
\indent For \textit{ANTs} registration with the affine and deformable component (SyN) there is good alignment on the middle slices but the top slice has some pixels missing, and the lower slices have the ocular orbits in multiple slices (which are not present in the atlas for these slices) indicating poor alignment in 3D. \textit{Demons} exhibits large deformations for all three slices and therefore, this tool is not ideal for clinical applications involving FLAIR MRI. \textit{SimpleElastix} seems to align the images over most spatial locations, except the lower slices as they contain  ocular orbits for slices that do not anatomically align with the atlas.  \textit{Voxelmorph} exhibits similar trends with good alignment in middle slices.  The lower slices however contain ocular orbits in slices that are not present in the atlas. The remaining rows show results for the proposed work.  First the results of only the affine component, \textit{FlowReg-A}, is displayed.  There is excellent alignment in the bottom and middle slices as well as in the top image slices indicating high anatomical alignment with the atlas. When combining \textit{FlowReg-A} with \textit{FlowReg-O} in the last row, the overall similarity and alignment with the atlas is improved. The shape of the head is more similar to that of the fixed volume slices, and also the top slice is more anatomically aligned.\\  
\begin{figure}
    \centering
    \includegraphics[width=\textwidth]{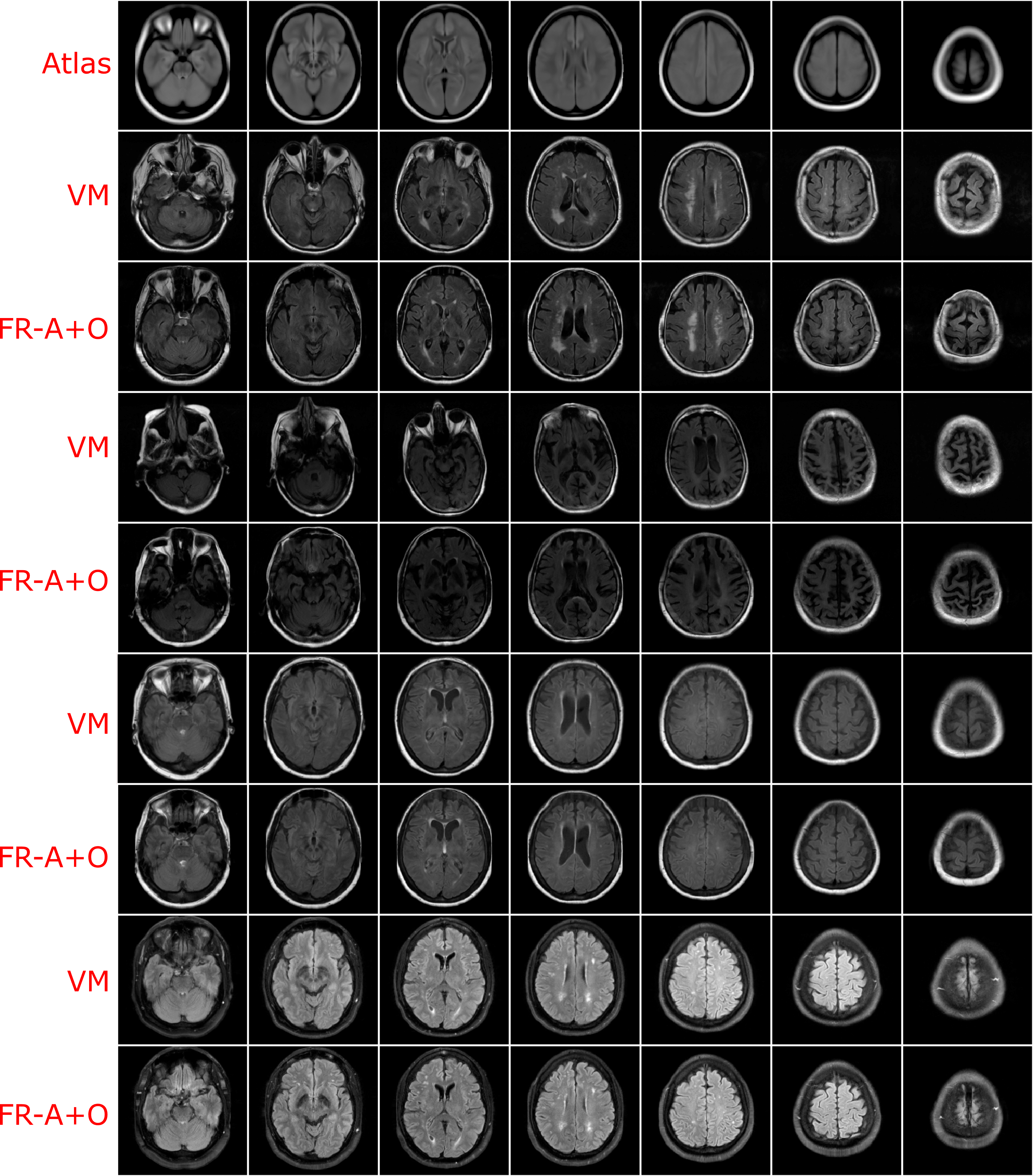}
    \caption{Visual comparison between Voxelmorph and FlowReg-A+O.  The top row contains the fixed volume $F(x,y,z)$ and the columns display sample registration results over four subjects for each method.}
    \label{fig:frao_vs_vm}
\end{figure}
\indent To examine the effect of anatomical alignment between the two CNN methods (\textit{VoxelMorph} and the proposed \textit{FlowReg}), see Figure \ref{fig:frao_vs_vm}. The top row has the fixed volume $F(x,y,z)$ and the remaining columns contain sequential registration results for four subjects using \textit{VoxelMorph} and \textit{FlowReg}. \textit{FlowReg} consistently provides a more accurate alignment of the slice data, and is more consistent in the anatomy it is representing across cases. This is especially evident in the bottom slices, where images registered with \textit{Flowreg} contain ocular orbits only in the correct slices, and in the top slices, where the top of the head is more accurate in size, shape and anatomy as compared to \textit{Voxelmorph}.\\
\indent To quantitatively compare performance across the methods, the average of the evaluations metrics from the testing set are shown in Table \ref{tab:metric_avgs}. The results for each category of validation metric will be described next. 
\begin{table}[]
\caption{Average registration performance (structural integrity, spatial alignment and intensity-based metrics). \textbf{Bold} values are best performance across methods while \underline{underline} is best among deep learning methods. $\simeq X$ indicates a value closest to X is best, $\downarrow$ lowest value is best, $\uparrow$ highest value is best. $\Delta PV_{vent}\times10{^-3}$, $\Delta PV_{wml} \times 10^{-3}$, $MAID \times 10^{-3}$, $MAID-zp \times 10^{-3}$.}
    \label{tab:metric_avgs}
    \centering
\begin{tabular}{@{}llcccccc@{}}
\toprule
&                   & ANTs   & Demons  & SE    & VM    & FlowReg-A      & FlowReg-A+O \\ \midrule
\multirow{6}{*}{\rotatebox{-90}{Structural}}&$\Delta V_{brain}\simeq1$ & 1.28   & 1.47    & 1.25  & \underline{\textbf{1.01}}  & 1.14  & 1.11 \\
& $\Delta V_{vent}\simeq1$  & 1.27   & 1.31    & 1.60  & 0.64  & \underline{\textbf{0.96}}  & 0.86 \\
& $\Delta V_{wml}\simeq1$   & 0.84   & \textbf{1.01}    & 0.81  & 0.35  & \underline{0.67}           & 0.53 \\
& $\Delta PV_{vent}\simeq0$ &\textbf{ 0.52} & -2.80  & 7.19  &-23.11   & \underline{-7.31} &-11.52 \\
& $\Delta PV_{wml}\simeq0$  & -5.17 &\textbf{-4.74}  &-5.67  &-20.17    & \underline{-7.14} &-11.20 \\
& $\Delta SSD\simeq0$       &\textbf{-4.70}  & -8.66   & -21.81 & 29.63 & 14.58       & \underline{13.82}\\ \midrule
\multirow{3}{*}{\rotatebox{-90}{Spatial}}& HA-$\varsigma\downarrow$     & 6.21   & 5.16    & \textbf{2.981}  & 10.83  & 6.89  & \underline{3.91} \\
& PWA-$\Sigma\downarrow$       & 1.24   & 2.40    & 1.92   & 1.47  & 1.15   & \underline{\textbf{0.65}} \\
& Brain-DSC$\uparrow$          & \textbf{0.88}     & 0.77      & 0.87      & 0.84      & \underline{0.86}     & 0.85 \\ \midrule
\multirow{4}{*}{\rotatebox{-90}{Intensity}}& MI$\uparrow$                 & 0.24   & 0.13    & 0.16   & 0.20  & 0.25   & \underline{\textbf{0.29}} \\
& R$\uparrow$                  & 0.64   & 0.41    & 0.39   & 0.60  & 0.65   & \underline{\textbf{0.80}} \\
& MAID$\simeq0$                & 5.24 & 6.26 & 5.99 & 5.54 & \underline{\textbf{5.08}}   & 5.33   \\
& MAID-zp$\simeq0$             & \textbf{0.53} & 1.99  & 1.35  & 1.16  & 0.86 & \underline{0.84} \\
\bottomrule
\end{tabular}
\end{table}
%
%
%
\subsubsection{Structural Integrity}
\begin{figure}[th!]
 \centering
     \includegraphics[width=0.6\textwidth]{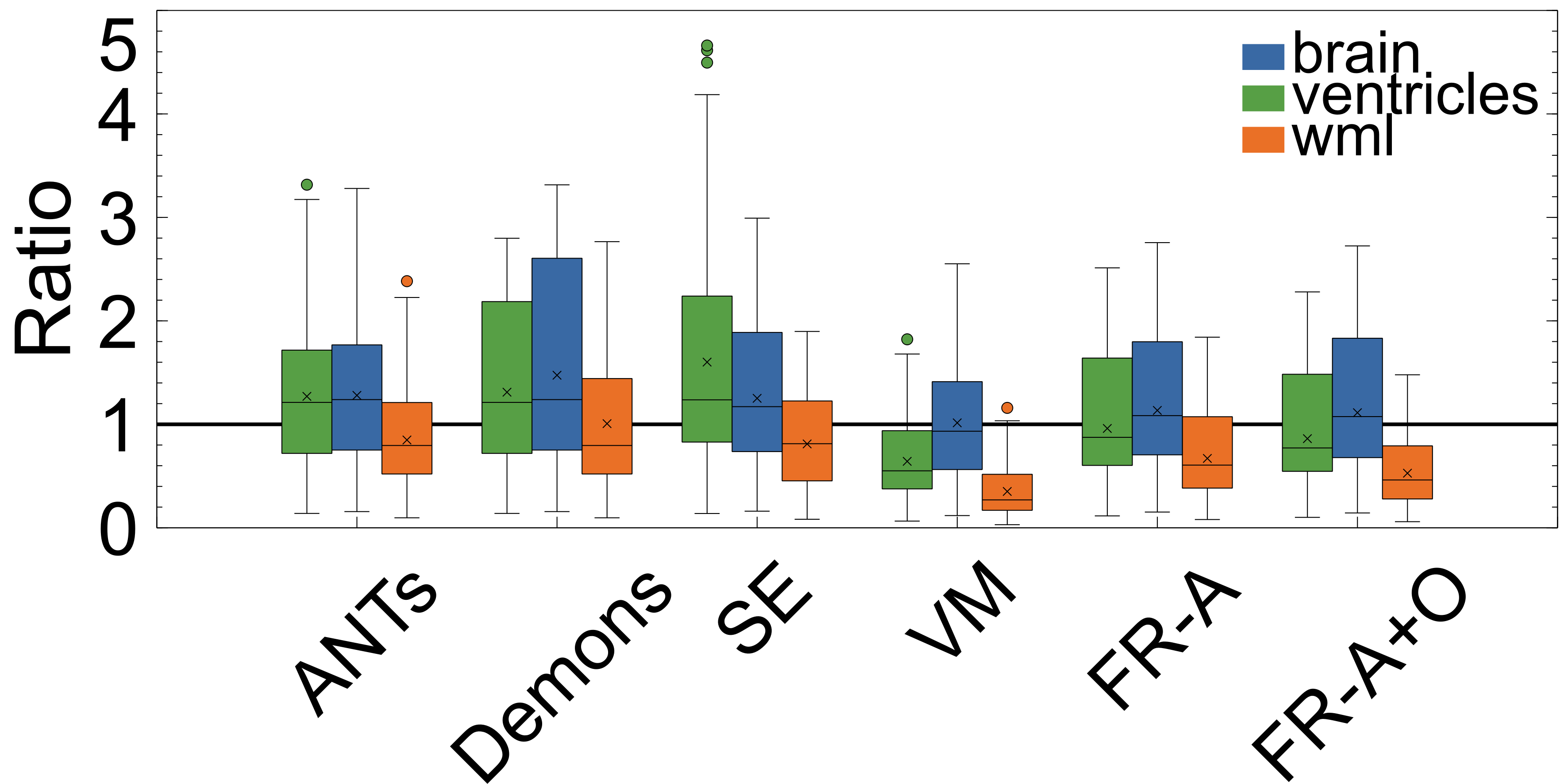}
     \caption{Structural volumetric ratio for brain, ventricles and WML.}
     \label{fig:vol_ratio}
 \end{figure}

\begin{figure}[h]
     \centering
     \includegraphics[width=.6\textwidth]{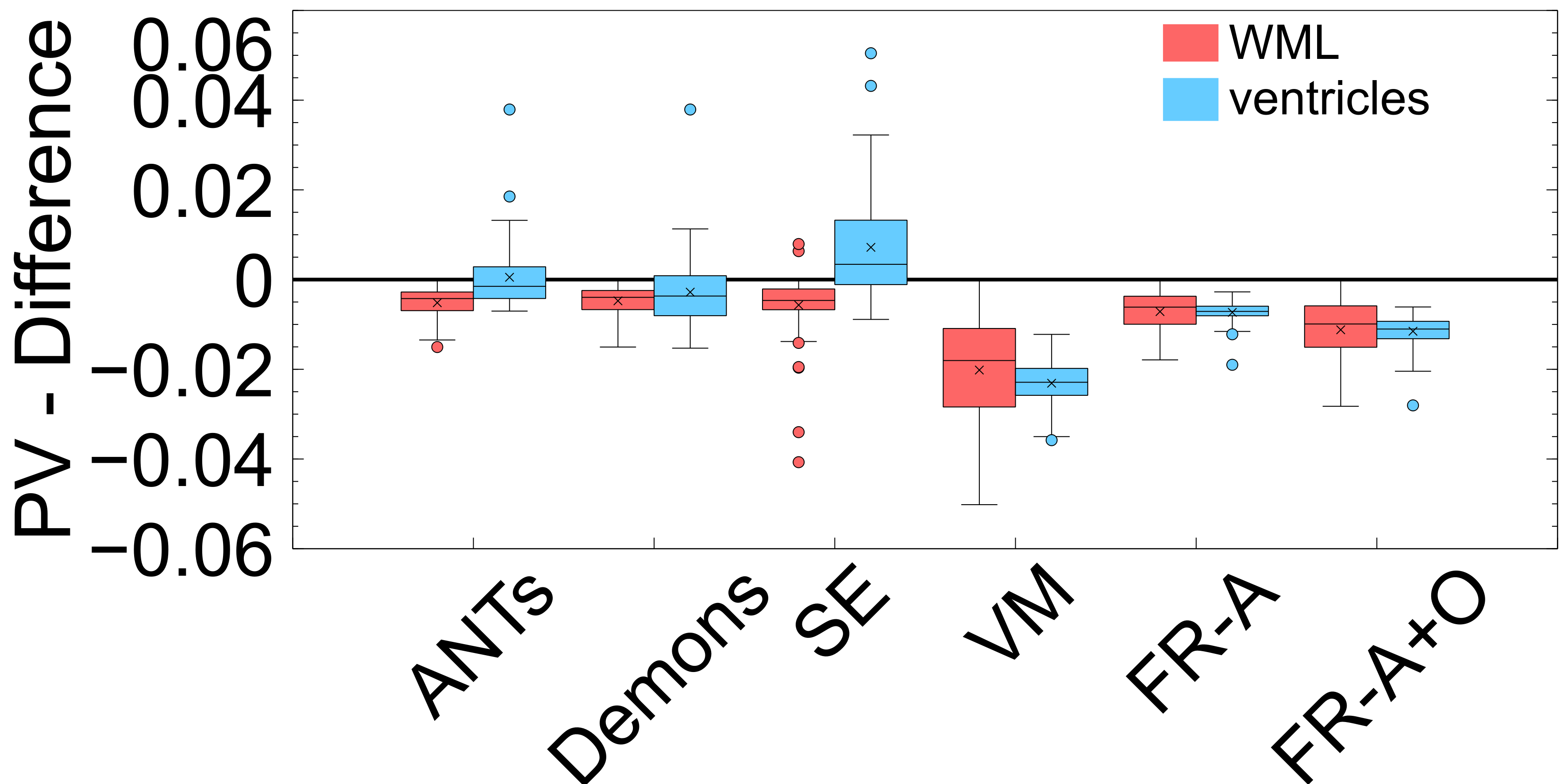}
     \caption{Proportional Volume (PV) difference of ventricles and WMLs.}
     \label{fig:pv}
 \end{figure}

\begin{figure}[h]
     \centering
     \includegraphics[width=0.6\textwidth]{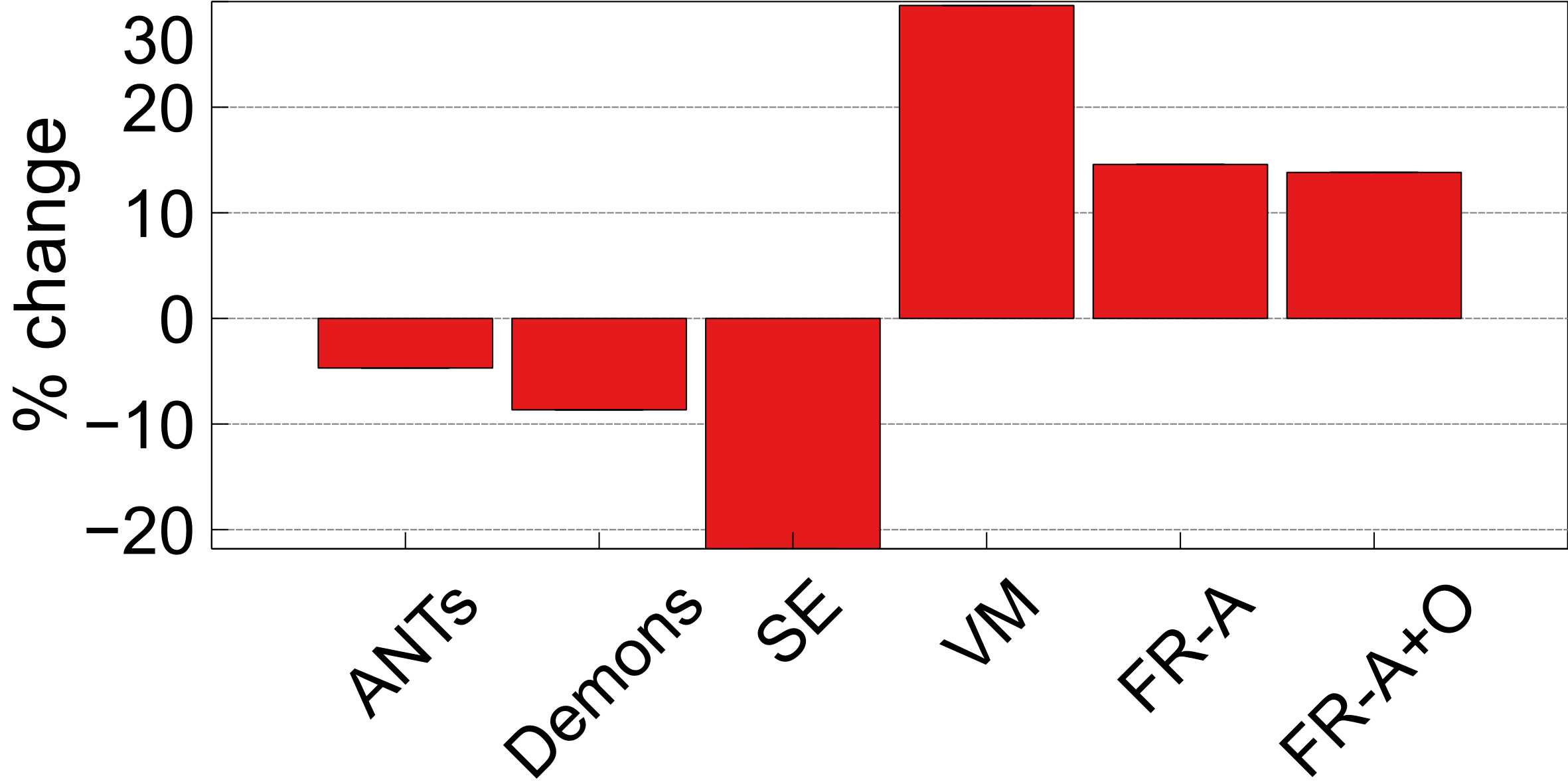}
     \caption{Surface to surface Distance (SSD) percent change of ventricles.}
     \label{fig:ssd}
 \end{figure}

To ensure structures of interest are not distorted and integrity is maintained, the following structural integrity metrics are examined: change in proportional volume ($\Delta PV$), the volume ratio ($\Delta V$), and the change in the structural similarity distance ($\Delta SSD$). The average of the metrics are listed in Table \ref{tab:metric_avgs} and the corresponding plots are shown in Figures \ref{fig:pv}, \ref{fig:vol_ratio}, and \ref{fig:ssd}.\\
%
%
\indent PV measures the proportional volumes of WMLs and ventricles before and after each registration. It is quantified as the PV change, $\Delta PV$, and the results are shown in Figure: \ref{fig:pv} while the average measures are in Table \ref{tab:metric_avgs}. The results nearest the zero line indicate the least change in PV compared to pre-registration and the least deformation. The PV metric shows that for all registration techniques the relative volumes of objects are mostly enlarged after deformation. The only cases where structures were decreased in size were the ventricles for the non-learning based methods \textit{ANTs} and \textit{SimpleElastix} for the ventricles. The least amount of distortion as quantified by the PV difference, for both ventricles and WMLs, is seen using the \textit{ANTs} and \textit{Demons} registration methods. The largest change in the WML and ventricles is seen in with \textit{VoxelMorph}. \textit{FlowReg-A} and \textit{FlowReg-O} slightly enlarge both ventricles and WML. \textit{FlowReg-A} has a PV difference for the ventricles of  $-7.3 \times 10^{-3}$ and a WML with  $-7.1 \times 10^{-3}$. \textit{FlowReg-A+O}, the combination of the affine and optical flow steps, shows approximately a $-11.5 \times 10^{-3}$ change in PV for the ventricles and $-11.2 \times 10^{-3}$ change in PV for the WML class. Since the two steps are performed in a sequential manner, the difference between \textit{FlowReg-A} and \textit{FlowReg-A+O} would provide the amount of change in PV provided by \textit{FlowReg-O}, which was found to be $4.2\times 10^{-3}$ and $4.1\times 10^{-3}$ for ventricles and WMLs, respectively. These values are closer to that of \textit{ANTs} registration.\\
\indent The volume ratio metrics quantify how much the structures of interest have decreased ($>1$) or increased in volume $(<1)$ after registration.  Ideally, structures would remain the same size after registration ($\Delta V_s = 1$).  As shown in Table \ref{tab:metric_avgs} and Figure \ref{fig:vol_ratio}, the volume ratio of the brain is most unchanged through registration with \textit{VoxelMorph}, followed closely by the proposed work (\textit{FlowReg-A} and the total pipeline \textit{FlowReg-A+O}). The ventricles are most similar to the original using \textit{FlowReg-A} and \textit{FlowReg-A+O}, where the size of the ventricles were increased slightly.  In contrast to traditional registration algorithms where ventricles mostly decrease in size, both \textit{FlowReg} and \textit{VoxelMorph} increase the size of the ventricles, where the size of the ventricles in \textit{VoxelMorph} have approximately doubled. In terms of WML, \textit{Demons} was the most favourable as the WML volume was almost  unchanged after registration.  This may be due to the fact this registration scheme seemed to mainly warp the boundary surrounding the head. Compared to the deep learning methods, in terms of WML enlargement, it seems that the traditional registration methods are more favourable in this regard, with \textit{FlowReg-A} providing the lowest volume increase out of all deep learning methods.\\ 
%
%
\indent The third integrity metric considered is SSD, which measures the shape of an anatomic object (this case the ventricles) with respect to the boundary of the brain.  To measure the extent to which the shape of the ventricles has changed in shape before and after registration, the difference, or percent change in SSD ($\Delta SSD$) is measured over the testing dataset and reported in Figure \ref{fig:ssd} for each of the registration methods. The lowest values are observed after registration using \textit{ANTs} and \textit{Demons}, followed by \textit{FlowReg-A+O}. \textit{FlowReg-A+O} has a change of around $13.8\%$ after registration which when compared to \textit{FlowReg-A}, the difference is about $1\%$ which is the assumed contribution from \textit{FlowReg-O} only. Since majority of the warping is done in 2D and largely affects the outer region of the brain and head, this metric exhibits that \textit{FlowReg-O} maintains the shape of the brain and ventricles. The highest structural change when measured with the SSD validation metric is noticed using \textit{SimpleElastix} and \textit{VoxelMorph}.  
%
%
\subsubsection{Spatial Alignment}
\begin{figure}
    \centering
    \includegraphics[width=\textwidth]{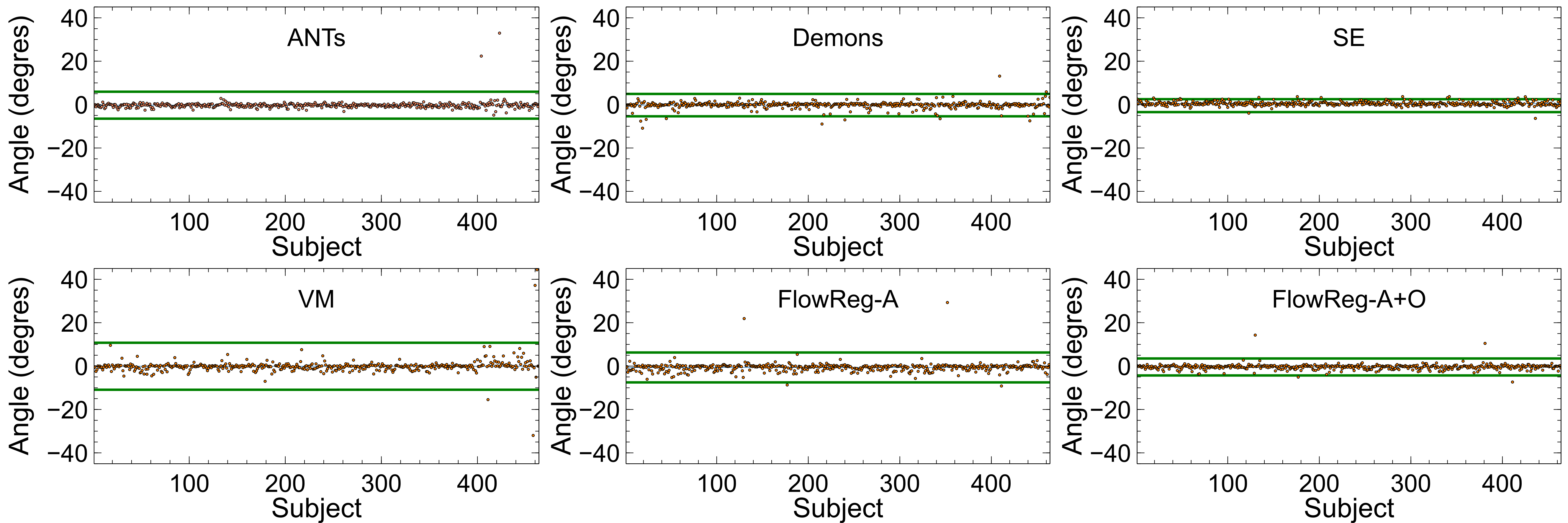}
    \caption{Head angle ($HA$) measure. Blue solid line indicates the average $\mu$ and the green line is the spread, $\sigma_{HA}$.}
    \label{fig:HA}
\end{figure}
\begin{figure}
    \centering
    \includegraphics[width=\textwidth]{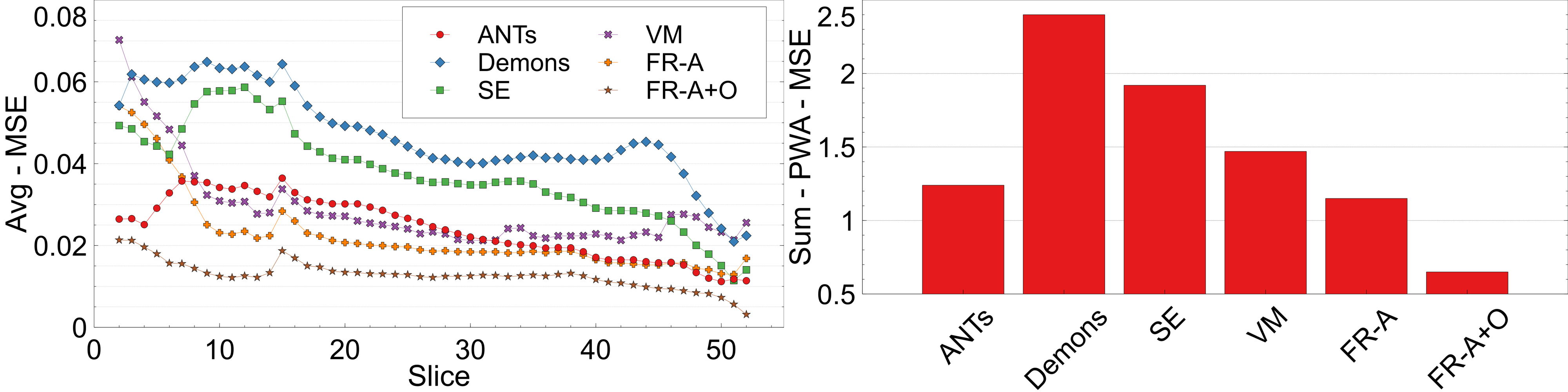}
    \caption{Pixelwise agreement (PWA). Left: the slice-wise alignment error $PWA(z)$ as a function of slice number $z$ in the registered dataset. Right: average PWA over all slices $\sum_z PWA(z)$.}
    \label{fig:pwa_mse}
\end{figure}
\begin{figure}
    \centering
    \includegraphics[width=0.5\textwidth]{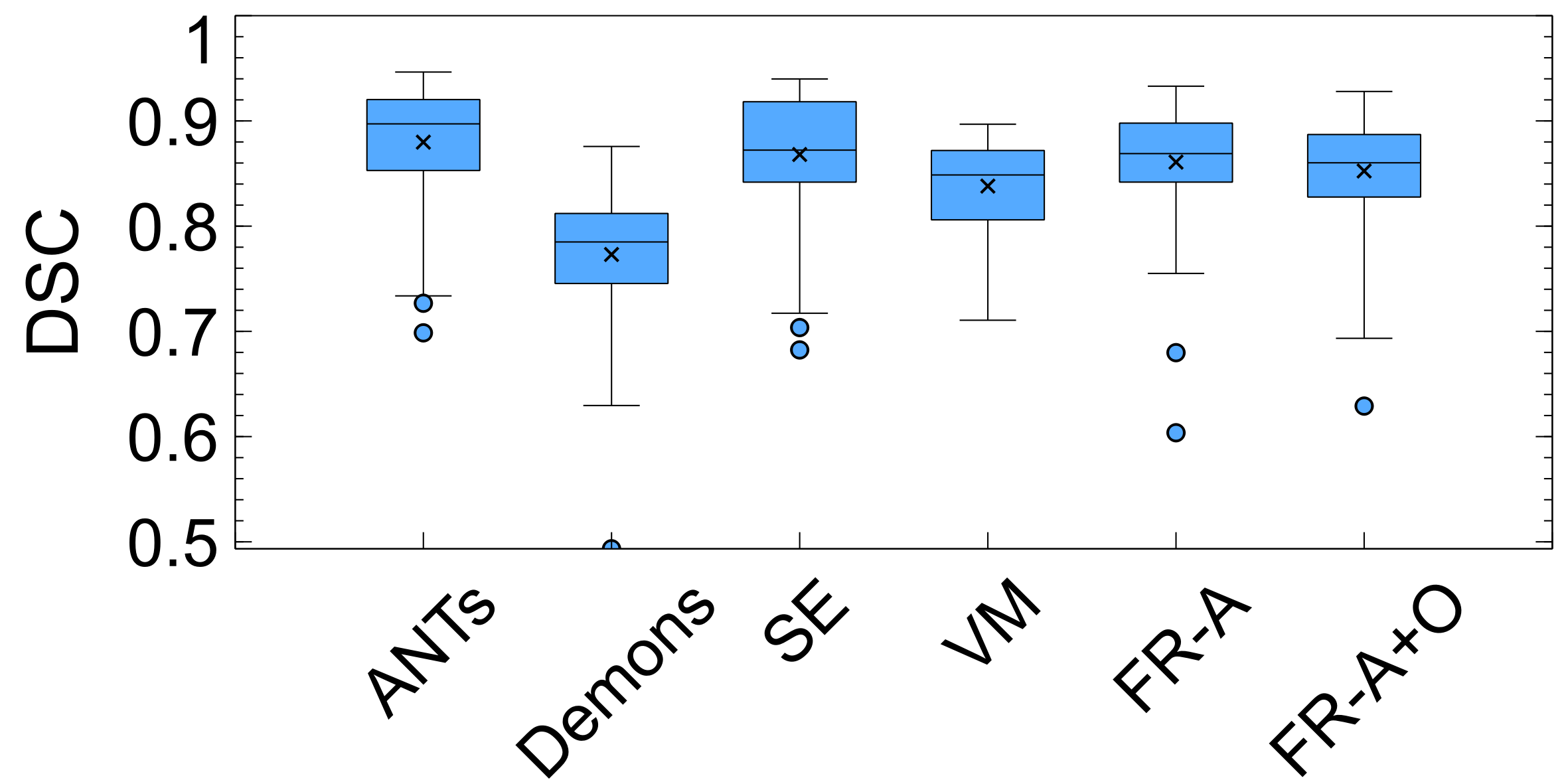}
    \caption{Dice Similarity Coefficient (DSC) between fixed and registered brain masks.}
    \label{fig:brain_dsc}
\end{figure}
\begin{figure}[!h]
    \centering
    \includegraphics[width=\textwidth]{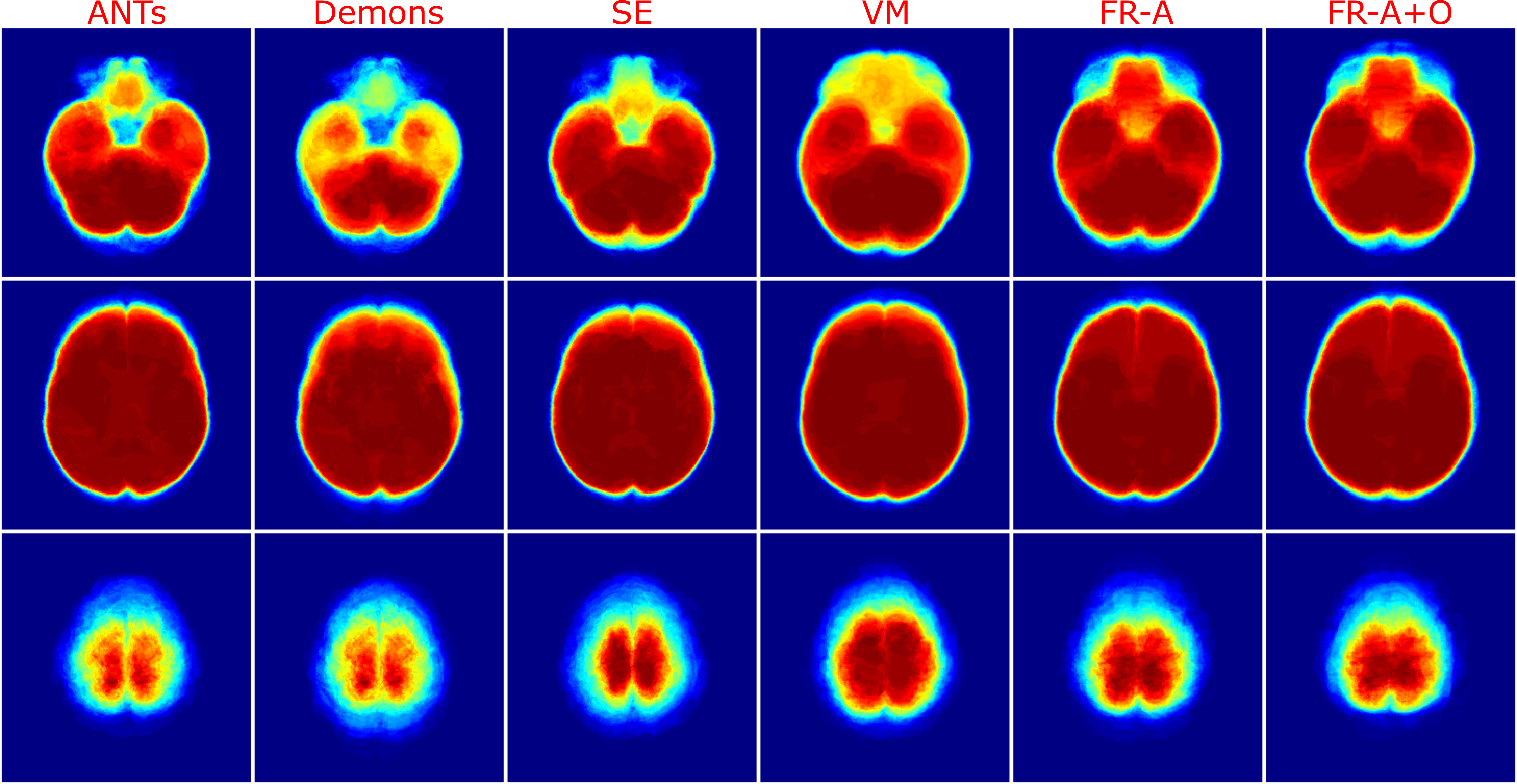}
    \caption{The average brain mask generated by each registration method. Red areas indicate high agreement, and blue indicates poor agreement.}
    \label{fig:brain_heatmaps}
\end{figure}

Figure \ref{fig:HA} displays the head angle (HA) results computed over all testing volumes. For best performance, the $HA$ would be ideally 0 (i.e. aligned with the midsagital plane), with minimal spread across the dataset to indicate consistency. The spread of the HA metric is denoted by $\varsigma_{HA}$ (average values can be found in Table \ref{tab:metric_avgs}) and is shown by the green line in Figure \ref{fig:HA} which indicates three standard deviations away from the mean. A tighter clustering around 0 degrees indicates less deviation from the midsagittal plane (or lower HA over the entire registered dataset). As seen, the lowest spread from the mean is seen by \textit{SimpleElastix} and \textit{FlowReg-A+O} registration methods, indicating these methods produce the most consistent spatial alignment with the midsaggital plane. It is also noted that the performance of \textit{FlowRegA+O} compared to the affine only \textit{FlowReg-A} volumes shows a reduction in the spread of the HA through the application of the optical flow algorithm, which indicates that \textit{FlowReg-O} improves overall alignment. The largest spread (or higher variability of the HA) is obtained by  \textit{Voxelmorph}.\\
\indent Pixelwise Agreement (PWA) is measured by calculating the per-slice mean-squared error (MSE) when compared to respective slices from the original atlas $F(x,y,z)$. A lower value  of PWA  indicates intensity and spatial alignment across slices in a registered dataset. PWA is computed on a slice-by-slice basis for slice $z$ by $PWA(z)$ which is summed over all slices in the volumes to get a volume-based PWA, $\sum_z PWA(z)$.  The slice- and volume-based PWA for the registered, testing dataset are reported in Fig. \ref{fig:pwa_mse}.  The lowest PWA over all slices is \textit{FlowReg-A+O} followed by \textit{FlowReg-A}, indicating there is maximal intensity and spatial alignment across slices using the proposed work. The highest error is seen by \textit{Demons}, \textit{SE}, and \textit{VoxelMorph}.\\
\indent The last spatial alignment measure investigated is the $DSC$ between the registered brain masks from the moving volumes, and the brain mask of the fixed atlas. Figure \ref{fig:brain_dsc} and Table \ref{tab:metric_avgs} contains the average $DSC$ values for each registration method over the testing dataset. The largest agreement is for \textit{ANTs}, \textit{SE}, \textit{FlowReg-A}, and \textit{FlowReg-A+O}.  The lowest spatial overlap comes from the \textit{Demons} method. To visualize spatial alignment, a heatmap is generated for each method by averaging the binary masks of the same slice in the registered output. Figure \ref{fig:brain_heatmaps} shows the heatmaps for a bottom, middle and top slice over all methods.  As can be seen, there is consistency in the lower slices for \textit{FlowReg}, as there is minimal ghosting artifacts in the heatmap.  However, with other methods, such as \textit{Demons} or \textit{VM}, there are many areas with inconsistencies in the posterior regions (likely where the ocular orbits occur).  In the middle slices, most methods seem to have good alignment, and the performance is somewhat comparable on the top slices.

\subsubsection{Intensity Alignment}
\begin{figure}
    \centering
    \includegraphics[width=\textwidth]{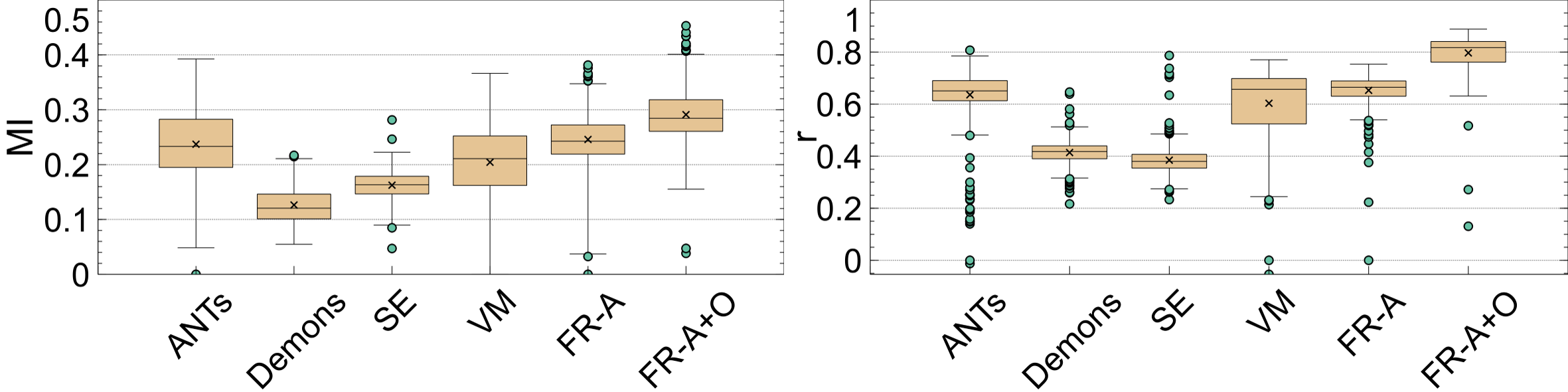}
    \caption{Intensity alignment validation over 464 testing volumes. Box and whisker plots. Left: mutual information. Right: correlation coefficient.}
    \label{fig:R_MI}
\end{figure}

\begin{figure}
    \centering
    \includegraphics[width=\textwidth]{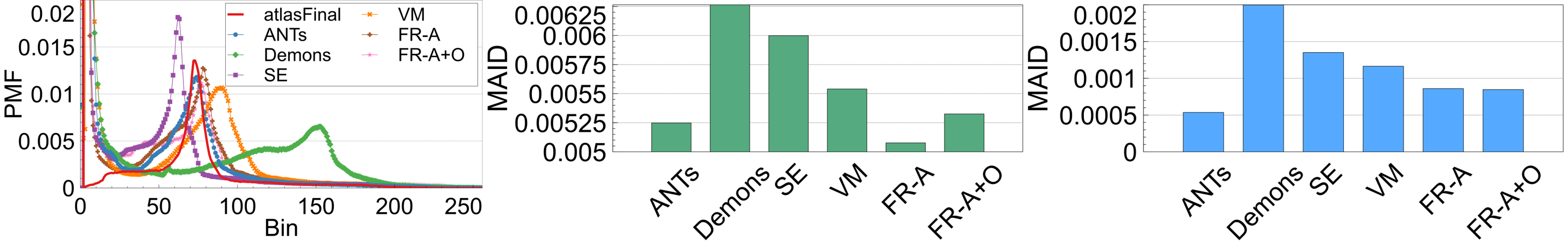}
    \caption{Left: Intensity distribution histograms of the atlases $A(x,y,z)$ created through registering all test volumes per method. Middle: MAID computed between intensity PMFs of the generated atlas and the original atlas (fixed). Right: MAID for PMFs with the background-nulled from bin $0$ to $20$ ($MAID_{zp}$)}
    \label{fig:pdfsmse}
\end{figure}

The last set of evaluation metrics investigated are the intensity alignment measures, and the average over the entire testing dataset is shown in Table \ref{tab:metric_avgs}.  Intensity alignment measures, mutual information (MI) and correlation (R), investigate how the probability mass functions of the registered volumes compare to the atlas' intensity distribution.  The intensity profiles in neuroimages are related to anatomy and pathology. Fig. \ref{fig:R_MI} shows boxplots of the MI and R metrics over the entire testing dataset. For both metrics, the highest MI and correlation are reported by \textit{FlowReg-A+O} and \textit{FlowReg-A} followed by \textit{ANTs}.  Therefore, the proposed work maintains and matches the intensity histograms the best over all competing methods.\\
%
%
%
%
%
%
\indent If a registration method generates images that have a high degree of spatial alignment, regions of high correspondence will have the same intensity characteristics reflected in the average.  If different tissue regions are being averaged, however, there will be mixing of neighbouring tissues and therefore, the intensity profile of the images will not be maintained. To measure this, registration-specific atlases are generated via synchronized averaging. The quality of these atlases $A(x,y,z)$ are quantitatively compared to the original template $F(x,y,z)$ by examining histogram differences via Mean Absolute Intensity Differences (MAID). In FLAIR MRI, histogram peaks represent significant tissue regions such as brain matter and cerebrospinal fluid (CSF). These peaks should be aligned in the newly generated atlas $A(x,y,z)$ with the original atlas $F(x,y,z)$. Fig. \ref{fig:pdfsmse} (left) shows the intensity histograms (normalized) of the atlases $A(x,y,z)$ compared to the histogram of the original fixed volume. It can be seen that the histogram of the \textit{FlowReg-O+A} and and \textit{FlowRegA} are very similar to that of the atlas for the middle (major) peak (which corresponds to the brain tissue). To quantitatively measure the similarity of histograms, the MAID is computed between the original and new generated atlases in Figure \ref{fig:pdfsmse} (middle) and the lowest error is found with \textit{FlowReg-A}. We analyze a second set of results for a thresholded histogram that removes the background noise peak from the histogram. This MAID, computed on the histogram without the noise peak is called  $MAID-zp$ and is shown in Figure \ref{fig:pdfsmse} (right). In these results, \textit{ANTs}, \textit{FlowReg-A}, and \textit{FlowReg-A+O} provide the best performance indicating good spatial and intensity alignment in the registered outputs for these methods. The performance of  \textit{FlowReg-A+O} and \textit{FlowReg-A} are similar, indicating \textit{FlowReg-O} does not distort the intensity histogram. The highest error is observed in \textit{Demons}. \\
%
%
%
%
%
%
%
\section{Discussion} \label{sec:discussion}
Table \ref{tab:metric_avgs} contains the summary of all the validation metrics. In comparison to all methods (\textit{ANTs}, \textit{Demons}, \textit{SE}, \textit{VM}), the proposed \textit{FlowReg} framework (\textit{FlowReg-A+O}) achieves the highest performance across the spatial alignment metric (PWA) which indicates excellent slice to slice correspondence between registered datasets and the fixed volume. This can be attributed to the initial alignment of the volumes in 3D using the affine component, followed by the slice-by-slice refinement  using optical flow in 2D that performs fine pixel movements. \textit{FlowReg} also achieves high intensity similarity based on the MI and R metrics, which indicates the histograms of the registered volumes and that of the atlas are aligned. The correlation loss function may contribute to this phenomena since it enforces global intensity similarity between images. Since intensity distributions are related to the tissue content in the images, if the histograms are more aligned in the registered images, it will make subsequent analysis consistent and comparable across patients. \textit{FlowReg-A} also was a top performer for several metrics, namely the volumetric ratio metric for the ventricles ($\Delta V_{vent}$) and the Mean Absolute Intensity Difference (MAID). The ventricular integrity metric indicates that the related shapes and volumes are maintained the best using \textit{FlowReg-A}. The intensity similarity can also be attributed to the correlation loss function in the affine network. Among the deep learning frameworks either \textit{FlowReg-A} or \textit{FlowReg-A+O} outperform \textit{Voxelmorph} in all metrics except for the structural integrity metric for the brain. This may be due to the deformation field calculated by \textit{Voxelmorph}, which was found to have lower vector magnitudes at the periphery of the head indicating little displacement in these regions. Since \textit{Voxelmorph} was trained and tested for other neuroimaging sequences (i.e. T1), perhaps this architecture is not suited for FLAIR neuroimages. \\
\indent Overall, \textit{FlowReg} maintains anatomical and structure features while obtaining high intensity similarity to the fixed volume and excellent spatial alignment across the testing datasets.  This can be attributed to several reasons. First, \textit{FlowReg-A} performs the affine alignment in 3D which will globally deform the volume to achieve maximal correspondence. Further, \textit{FlowReg-O} calculates the 2D displacement field and refines the movement of pixels on a slice-by-slice basis. The optical flow model architecture has been adapted from the video processing field to medical images. The advantage of this approach is that it is able to calculate small differences and perform the refinements needed to obtain correspondence. The three component loss function (photometric, smoothness, and correlation) perform three separate but important roles. The photometric loss, which is a pixelwise difference, ensures that pixels with similar intensities are displaced to areas of similar intensities and is the refinement component. The role of the smoothness loss is to ensure that for the calculated optical flow, continuity of the flow fields is encouraged. The correlation loss operates on the overall histogram intensity alignment between the moving and fixed volumes. Finally, the Charbonnier penalty function was used to reduce the effect of gross outliers to ensure the structural integrity of anatomy and pathology was maintained. It can be seen that with the combination of these loss functions, \textit{FlowReg-A+O} performs the best for the intensity measures, MI, R, and for the spatial alignment measure PWA and is likely due to the complementary nature of the loss components. As the deformation vector field is generated for each pixel in the moving image, the photometric loss specifically minimizes the difference between similar intensity pixels in the moving and fixed images and can therefore regress accurate vectors for pixel displacement. Similarly, the correlation loss component operates on the global intensity differences and contributes to accurate global flow regression. \\
\indent To investigate the relative run-time speed for each algorithm, each method was evaluated on ten randomly sampled volumes from the testing set. The average computation times per method for these ten volumes are shown in Table \ref{tab:runtime}. Note that the times reported for \textit{FlowReg} are for the two components \textit{FlowReg-A} and \textit{FlowReg-O} separately.  For \textit{Voxelmorph}, only the test time for the CNN network is shown. As can be seen, the fastest algorithm is \textit{FlowReg-A} with an average time of $1.72$ seconds per 3D volume, followed by \textit{Voxelmorph} at 4.31s and then \textit{FlowReg-O} with 6.9s.  Given the testing time for \textit{FlowReg-A} and \textit{FlowReg-O}, the total time to register a volume (3D) and all the slices (2D) is 8.62s. Compared to the traditional approaches, the proposed method is faster by 2.6-4.7$\times$. It is to be noted that the time reported for \textit{Voxelmorph} is only the CNN network testing time, but for optimal performance affine transformation is required first using tools such as ANTs which adds additional computation time.  Further, all deep learning methods outperform the traditional iterative methods by a large margin, indicating that the transferable nature of CNN-based registration tools are efficient and effective.  As with many CNN-based systems, the challenge is the training time; the average training time on the 3714 training volumes for FlowReg-A was about 18 hours, while FlowReg-O was just over 5 days.  As training can be completed offline, it is not too big of a concern for real-time applications.\\
\begin{table}[]
    \centering
    \caption{Average run-time for ten testing 3D volumes for each registration method.}
    \label{tab:runtime}
    \begin{tabular}{@{}lllllll@{}}
    \toprule
         & ANTs & Demons & SE & VM-only & FlowReg-A &FlowReg-O  \\ \midrule
         & 27.81s & 22.77s & 40.33s & 4.31s & \underline{\textbf{1.72s}} & 6.90s \\ \bottomrule
    \end{tabular}
\end{table}
\indent Another interesting observation is that for WML structural integrity, all CNN-based solutions perform poorly compared to classic iterative-based approaches. Hyperintense lesions are usually the brightest spots in a brain MRI, thus when performing an alignment to a template (such as an averaged atlas) these hyperintense regions will be represented as areas of high dissimilarities. A network with a loss function that attempts to mitigate pixel differences between the two volumes will attempt to over-correct these areas and displace the pixels in undesirable ways. This displacement will change the registered volume's WMLs from a structural and volumetric perspective. A possible solution that can be investigated to mitigate this problem in future works is lesion-inpainting \citep{sdika2009nonrigid} prior to registration. Inpainting frameworks mask the lesions with the average intensities of the surrounding brain tissue. However, this approach would require either manual or automatic segmentation of WMLs which is an active area of research. Additionally, future works could also consider the incorporation of skull-stripping as a preprocessing step, to remove all non-cerebral tissues. This could improve performance since a lot the warping is occurring in regions where pixel intensities are high, which correspond to extra-cranial structures. If these areas are removed it is possible that more of the pixel movement would be focused to resolve areas of higher differences in the brain.\\
\indent When considering all validation groups and metrics, the traditional iterative based registration methods perform well over many of the structural integrity metrics such as $\Delta V_{wml}$, $\Delta PV_{vent}$, $\Delta PV_{wml}$, indicating that these methods do not deform the WML and ventricles as much as the CNN-based methods.  In terms of \textit{Demons}, although it does perform the best according to the $\Delta V$ and $\Delta PV$ metrics, when visually examining several volumes as depicted in Figure \ref{fig:vol_montage}, the cerebral tissue seems warped in a manner that is uncharacteristic of FLAIR MRI. Further, structural changes are visible around the edges of the sulci and the WML themselves have been smeared and blended with the remainder of the gray matter of the brain. A limitation of the design is noted in the \textit{FlowReg-A} registration when it comes to the $\Delta V$ and $\Delta PV$. As \textit{FlowReg-A} operates on the volume using an affine matrix, which is a global transformation that equally warps every voxel in the volume, the proportional and volumetric difference of a structure before and after registration should remain unchanged. In our experiments we did notice differences for \textit{FlowReg-A} and reported the differences. This outcome is likely due to slice thickness, limited pixel resolution, and the slice gap of $~3mm$ during image acquisition.  \textit{ANTs} maintains moderate distortion over all structural integrity metrics and the best for the ventricular metrics. One reason for this could be due to the Symmetric Normalization, where both the moving and the fixed volumes are warped symmetrically to a similarity "mid-point" \citep{avants2008symmetric}. \textit{SimpleElastix}, another iterative-based registration method, performs well for the Head Angle metric likely due to the two step process of an affine alignment followed by a B-spline transform \citep{klein2010elastix}. One major downside of using iterative-based methods for image registration is the lengthy computation for 3D neuroimaging volumes (as is seen in Table \ref{tab:metric_avgs}) and the lack of transferring this knowledge to new image pairs. \\
\indent Medical image registration is a preprocessing tool that can map two images to the same geometric space. Once images are registered, direct spatial comparisons can be made between the two images to quantify disease progression, treatment efficacy, pathology changes, and age-specific anatomical changes. The proposed \textit{FlowReg} model is able to warp a moving image to a fixed image space in an unsupervised manner, and is computed in a two-phase approach: initially for gross alignment in 3D, followed by fine-tuning in 2D on an image-by-image basis which is a novel approach. Alongside the registration framework, several clinically relevant validation metrics are proposed that we hope will be used by researchers in the future. 
\section{Conclusion} \label{sec:conclusion}
In this work we propose \textit{FlowReg}, a deep learning-based framework that performs unsupervised image registration for multicentre FLAIR MRI. The system is composed of two architectures: \textit{FlowReg-A} which affinely corrects for gross differences between moving and fixed volumes in 3D followed by \textit{FlowReg-O} which performs pixelwise deformations on a slice-by-slice basis for fine tuning in 2D. Using 464 testing volumes, with 70 of the imaging volumes having ground truth manual delineations for ventricles and lesions, the proposed method was compared to \textit{ANTs}, \textit{Demons}, \textit{SE}, and \textit{Voxelmorph}. To quantitatively assess the performance of the registration tools, several proposed validation metrics were used. These metrics focused on structural integrity of tissues, spatial alignment, and intensity similarity. Tissue integrity was analyzed using volumetric and structural measures: rolumetric ratio, proportional volume (PV), and structural similarity distance (SSD). Spatial alignment was analyzed with a Head Angle with respect to the saggital plane, Pixelwise Agreement, and Brain DSC. The intensity metrics measured the similarity in intensities and intensity distrubitons of the moving and fixed volumes with Mutual Information (MI), correlation (R), and Mean Intensity Difference. 

Experimental results show \textit{FlowReg} (\textit{FlowReg-A+O}) performs better than iterative-based registration algorithms for intensity and spatial alignment metrics, indicating that \textit{FlowReg} delivers optimal intensity and spatial alignment between moving and fixed volumes. Among the deep learning frameworks evaluated, \textit{FlowReg-A} or \textit{FlowReg-A+O} provided the highest performance over all but one of the metrics.  In terms of structural integrity metrics, \textit{FlowReg} provided moderate (or best) performance for the brain, ventricle and WML objects. The success of the proposed work can be attributed to: 1) the two-step registration process that consists of affine followed by optical flow deformations and 2) the three component loss function in optical flow that encourages global intensity similarity, while minimizing large deformations. Finally, the novel validation metrics to assess medical image registration provide the necessary context when compared to other registration methods.


\acks{We acknowledge the support of the Natural Sciences and Engineering Research Council of Canada (NSERC) through the NSERC Discovery Grant program.

We would also like to acknowledge the research mentorship received for this work from Dr. Konstantinos Derpanis (Ryerson University, Computer Science Dept.) and Jason J. Yu (York University, Computer Science Dept.) which included methodological input on the optical flow network and on the use of the penalty function, in addition to editing of the  manuscript.

The Canadian Atherosclerosis Imaging Network (CAIN) was established through funding from a Canadian Institutes of Health Research Team Grant for Clinical Research Initiatives (CIHR-CRI 88057). Funding for the infrastructure was received from the Canada Foundation for Innovation (CFI-CAIN 20099), with matching funds provided by the governments of Alberta, Ontario, and Quebec.
	
Data collection and sharing for this project was partially funded by the Alzheimer's Disease Neuroimaging Initiative (ADNI) (National Institutes of Health Grant U01 AG024904) and DOD ADNI (Department of Defence award number W81XWH-12-2-0012).}
%
\ethics{The work follows appropriate ethical standards in conducting research and writing the manuscript, following applicable laws and regulations.}
\coi{There are no conflicts of interest.}
\bibliography{sample}


\appendix 
\section*{Appendix A}
\subsection*{Imaging Dataset Details}
A summary of the datasets used, and demographical information are shown in Table \ref{tab:demographics}. The ADNI database consisted of images from three MR scanner manufacturers: General Electric ($n=1075$), Phillips Medical Systems ($n=848$), and Siemens ($n=2076$) with $18$ different models in total. In CAIN there is five different models across three vendors with General Electric ($n=181$), Phillips Medical Systems ($n=230$), and Siemens ($n=289$). The number of cases per scanner model and vendor are shown in Table \ref{tab:scanners} along with the ranges of the acquisition parameters. As can be seen, this dataset represents a diverse multicentre dataset, with varying scanners, diseases, voxel resolutions, imaging acquisition parameters and pixel resolutions. Therefore, this dataset will give insight into how each registration method can generalize in multicentre data.  

\begin{table*}[htpb]
\caption{Experimental datasets used in this work (CAIN and ADNI).}
\centering
\begin{tabular}{@{}lllllll@{}}
\toprule
Dataset & \# Subjects & \# Volumes & \# Slices & \# Centers & Age            & Sex F/M ($\%$)   \\ \midrule
CAIN    & 400         & 700        & 31,500    & 9          & $73.87\pm8.29$ & 38.0/58.6        \\
ADNI    & 900         & 4263       & 250,00    & 60         & $73.48\pm7.37$ & 46.5/53.5        \\ \bottomrule
\end{tabular}
\label{tab:demographics}
\end{table*}
\begin{table}[!h]
\caption{Detailed acquisition parameters and information for each dataset. }
    \label{tab:scanners}
    \centering
    \begin{adjustbox}{angle=-90}
    \begin{tabular}{@{}llllllllll@{}}
    \toprule
Dataset & Vendor                                                           & Model                                                        & \begin{tabular}[c]{@{}l@{}}TR\\ ($ms$)\end{tabular}       & \begin{tabular}[c]{@{}l@{}}TE\\ ($ms$)\end{tabular}          & \begin{tabular}[c]{@{}l@{}}TI\\ ($ms$)\end{tabular}       & \begin{tabular}[c]{@{}l@{}}Magnetic \\ Field ($B$)\end{tabular} & \begin{tabular}[c]{@{}l@{}}Pixel Size\\ ($mm^2$)\end{tabular} & \begin{tabular}[c]{@{}l@{}}Slice \\ Thickness\\ ($mm$)\end{tabular} & N    \\ \midrule
ADNI    & \begin{tabular}[c]{@{}l@{}}GE Medical \\ Systems\end{tabular}          & \begin{tabular}[c]{@{}l@{}}Discovery \\ MR750/w\end{tabular} & 11000                                                   & \begin{tabular}[c]{@{}l@{}}149.42 - \\ 153.13\end{tabular} & 2250                                                    & 3                                                             & 0.7386                                                                       & 5                                                                 & 614  \\
        &                                                                        & Signa HDxt                                                   & \begin{tabular}[c]{@{}l@{}}10002 -\\ 11002\end{tabular} & \begin{tabular}[c]{@{}l@{}}149.10 - \\ 192.65\end{tabular} & \begin{tabular}[c]{@{}l@{}}2200 - \\ 2250\end{tabular}  & 1.5 - 3                                                       & 0.7386 - 0.8789                                                              & 5 - 6                                                             & 461  \\ \cmidrule(l){2-10} 
        & \begin{tabular}[c]{@{}l@{}}Phillips\\ Medical\\ Systems\end{tabular}   & Ingenia                                                      & 9000                                                    & 90                                                         & 2500                                                    & 3                                                             & 0.7386                                                                       & 5                                                                 & 83   \\
        &                                                                        & Achieva                                                      & 9000                                                    & 90                                                         & 2500                                                    & 3                                                             & 0.6104 - 0.7386                                                              & 5                                                                 & 520  \\
        &                                                                        & Gemini                                                       & 9000                                                    & 90                                                         & 2500                                                    & 3                                                             & 0.7386                                                                       & 5                                                                 & 35   \\
        &                                                                        & Ingenuity                                                    & 9000                                                    & 90                                                         & 2500                                                    & 3                                                             & 0.7386                                                                       & 5                                                                 & 19   \\
        &                                                                        & Intera                                                       & \begin{tabular}[c]{@{}l@{}}6000 -\\ 9000\end{tabular}   & \begin{tabular}[c]{@{}l@{}}90 - \\ 140\end{tabular}        & \begin{tabular}[c]{@{}l@{}}2000 -  \\ 2500\end{tabular} & 1.5 - 3                                                       & 0.7386 - 0.8789                                                              & 5                                                                 & 191  \\ \cmidrule(l){2-10} 
        & Siemens                                                                & \begin{tabular}[c]{@{}l@{}}Biograph \\ mMR\end{tabular}      & 9000                                                    & 91                                                         & 2500                                                    & 3                                                             & 0.7386                                                                       & 5                                                                 & 13   \\
        &                                                                        & Prisma                                                       & 9000                                                    & 91                                                         & 2500                                                    & 3                                                             & 0.7386                                                                       & 5                                                                 & 5    \\
        &                                                                        & Skyra                                                        & 9000                                                    & 91                                                         & 2500                                                    & 3                                                             & 0.7386                                                                       & 5                                                                 & 213  \\
        &                                                                        & SymphonyTim                                                  & 10000                                                   & 125                                                        & 2200                                                    & 1.5                                                           & 0.8789                                                                       & 5                                                                 & 2    \\
        &                                                                        & TrioTim                                                      & \begin{tabular}[c]{@{}l@{}}9000 - \\ 11000\end{tabular} & \begin{tabular}[c]{@{}l@{}}90 - \\ 149\end{tabular}        & \begin{tabular}[c]{@{}l@{}}2250 - \\ 2800\end{tabular}  & 3                                                             & 0.7386 - 1                                                                   & 2 - 5                                                             & 1332 \\
        &                                                                        & Verio                                                        & 9000                                                    & 91                                                         & 2500                                                    & 3                                                             & 0.7386 - 1.0315                                                              & 5                                                                 & 511  \\ \midrule
CAIN    & \begin{tabular}[c]{@{}l@{}}GE Medical \\ Systems\end{tabular}          & \begin{tabular}[c]{@{}l@{}}Discovery \\ MR750\end{tabular}   & \begin{tabular}[c]{@{}l@{}}8000 - \\ 9995\end{tabular}  & \begin{tabular}[c]{@{}l@{}}140.84 - \\ 150.24\end{tabular} & \begin{tabular}[c]{@{}l@{}}2200 - \\ 2390\end{tabular}  & 3                                                             & 0.7386 - 0.8789                                                              & 3                                                                 & 181  \\ \cmidrule(l){2-10} 
        & \begin{tabular}[c]{@{}l@{}}Phillips \\ Medical \\ Systems\end{tabular} & Achieva                                                      & \begin{tabular}[c]{@{}l@{}}9000 - \\ 11000\end{tabular} & 125                                                        & 2800                                                    & 3                                                             & 0.1837                                                                       & 3                                                                 & 230  \\ \cmidrule(l){2-10} 
        & Siemens                                                                & InteraMR                                                     & 9000                                                    & 119                                                        & 2500                                                    & 3                                                             & 1                                                                            & 3                                                                 & 14   \\
        &                                                                        & Skyra                                                        & 9000                                                    & 119                                                        & 2500                                                    & 3                                                             & 1                                                                            & 3                                                                 & 162  \\
        &                                                                        & TrioTim                                                      & 9000                                                    & \begin{tabular}[c]{@{}l@{}}117 - \\ 122\end{tabular}       & 2500                                                    & 3                                                             & 1                                                                            & 3                                                                 & 113  \\ \bottomrule
\end{tabular}
\end{adjustbox}
\end{table}
\vskip .2in

\subsection*{Supplemental Tables and Figures}
\begin{table}[!h]
\centering
\caption{\textit{FlowReg-A} model, details of architecture in Figure \ref{fig:FlowReg-A}.}
\begin{tabular}{lllll}
\toprule
Layer       & Filters               & Kernel                & Stride                & Activation            \\
\midrule
fixedInput  & \multicolumn{1}{c}{-} & \multicolumn{1}{c}{-} & \multicolumn{1}{c}{-} & \multicolumn{1}{c}{-} \\
movingInput & \multicolumn{1}{c}{-} & \multicolumn{1}{c}{-} & \multicolumn{1}{c}{-} & \multicolumn{1}{c}{-} \\
concatenate & \multicolumn{1}{c}{-} & \multicolumn{1}{c}{-} & \multicolumn{1}{c}{-} & \multicolumn{1}{c}{-} \\
conv3D      & 16                    & 7x7x7                 & 2, 2, 1               & ReLu                  \\
conv3D      & 32                    & 5x5x5                 & 2, 2, 1               & ReLu                  \\
conv3D      & 64                    & 3x3x3                 & 2, 2, 2               & ReLu                  \\
conv3D      & 128                   & 3x3x3                 & 2, 2, 2               & ReLu                  \\
conv3D      & 256                   & 3x3x3                 & 2, 2, 2               & ReLu                  \\
conv3D      & 512                   & 3x3x3                 & 2, 2, 2               & ReLu                  \\
flatten     & \multicolumn{1}{c}{-} & \multicolumn{1}{c}{-} & \multicolumn{1}{c}{-} & \multicolumn{1}{c}{-} \\
dense       & 12                    & \multicolumn{1}{c}{-} & \multicolumn{1}{c}{-} & Linear               
\\ \bottomrule
\end{tabular}
\label{tab:flowreg-a}
\end{table}
\begin{table}[!h]
    \centering
    \caption{\textit{FlowReg-O} model details of architecture in Figure \ref{fig:FlowReg-O}.}
    \begin{tabular}{lllll}
    \toprule
Layer       & Filters               & Kernel                & Strides               & Activation            \\
    \midrule
fixedInput  & \multicolumn{1}{c}{-} & \multicolumn{1}{c}{-} & \multicolumn{1}{c}{-} & \multicolumn{1}{c}{-} \\
movingInput & \multicolumn{1}{c}{-} & \multicolumn{1}{c}{-} & \multicolumn{1}{c}{-} & \multicolumn{1}{c}{-} \\
concatenate & \multicolumn{1}{c}{-} & \multicolumn{1}{c}{-} & \multicolumn{1}{c}{-} & \multicolumn{1}{c}{-} \\
conv2D      & 64                    & 7x7                   & 2, 2                  & L-ReLu                \\
conv2D      & 128                   & 5x5                   & 2, 2                  & L-ReLu                \\
conv2D      & 256                   & 5x5                   & 2, 2                  & L-ReLu                \\
conv2D      & 256                   & 3x3                   & 1, 1                  & L-ReLu                \\
conv2D      & 512                   & 3x3                   & 2, 2                  & L-ReLu                \\
conv2D      & 512                   & 3x3                   & 1, 1                  & L-ReLu                \\
conv2D      & 512                   & 3x3                   & 2, 2                  & L-ReLu                \\
conv2D      & 512                   & 3x3                   & 1, 1                  & L-ReLu                \\
conv2D      & 1024                  & 3x3                   & 2, 2                  & L-ReLu                \\
conv2D      & 1024                  & 3x3                   & 1, 1                  & L-ReLu                \\
conv2D      & 2                     & 3x3                   & 1, 1                  & \multicolumn{1}{c}{-} \\
upconv2D    & 2                     & 4x4                   & 2, 2                  & \multicolumn{1}{c}{-} \\
upconv2D    & 512                   & 4x4                   & 2, 2                  & L-ReLu                \\
conv2D      & 2                     & 3x3                   & 1, 1                  & \multicolumn{1}{c}{-} \\
upconv2D    & 2                     & 4x4                   & 2, 2                  & \multicolumn{1}{c}{-} \\
upconv2D    & 256                   & 4x4                   & 2, 2                  & L-ReLu                \\
conv2D      & 2                     & 3x3                   & 1, 1                  & \multicolumn{1}{c}{-} \\
upconv2D    & 2                     & 4x4                   & 2, 2                  & \multicolumn{1}{c}{-} \\
upconv2D    & 128                   & 4x4                   & 2, 2                  & L-ReLu                \\
conv2D      & 2                     & 3x3                   & 1, 1                  & \multicolumn{1}{c}{-} \\
upconv2D    & 2                     & 4x4                   & 2, 2                  & \multicolumn{1}{c}{-} \\
upconv2D    & 64                    & 4x4                   & 2, 2                  & L-ReLu                \\
conv2D      & 2                     & 3x3                   & 1, 1                  & \multicolumn{1}{c}{-} \\
upconv2D    & 2                     & 4x4                   & 2, 2                  & \multicolumn{1}{c}{-} \\
upconv2D    & 32                    & 4x4                   & 2, 2                  & L-ReLu                \\
conv2D      & 2                     & 3x3                   & 1, 1                  & \multicolumn{1}{c}{-} \\
upconv2D    & 2                     & 4x4                   & 2, 2                  & \multicolumn{1}{c}{-} \\
upconv2D    & 16                    & 4x4                   & 2, 2                  & L-ReLu                \\
conv2D      & 2                     & 3x3                   & 2, 2                  & \multicolumn{1}{c}{-} \\
resampler   & \multicolumn{1}{c}{-} & \multicolumn{1}{c}{-} & \multicolumn{1}{c}{-} & \multicolumn{1}{c}{-} \\
\bottomrule
\end{tabular}
    \label{tab:flowreg-o}
\end{table}

\begin{figure}[h!]
    \centering
    \includegraphics[width=0.5\textwidth]{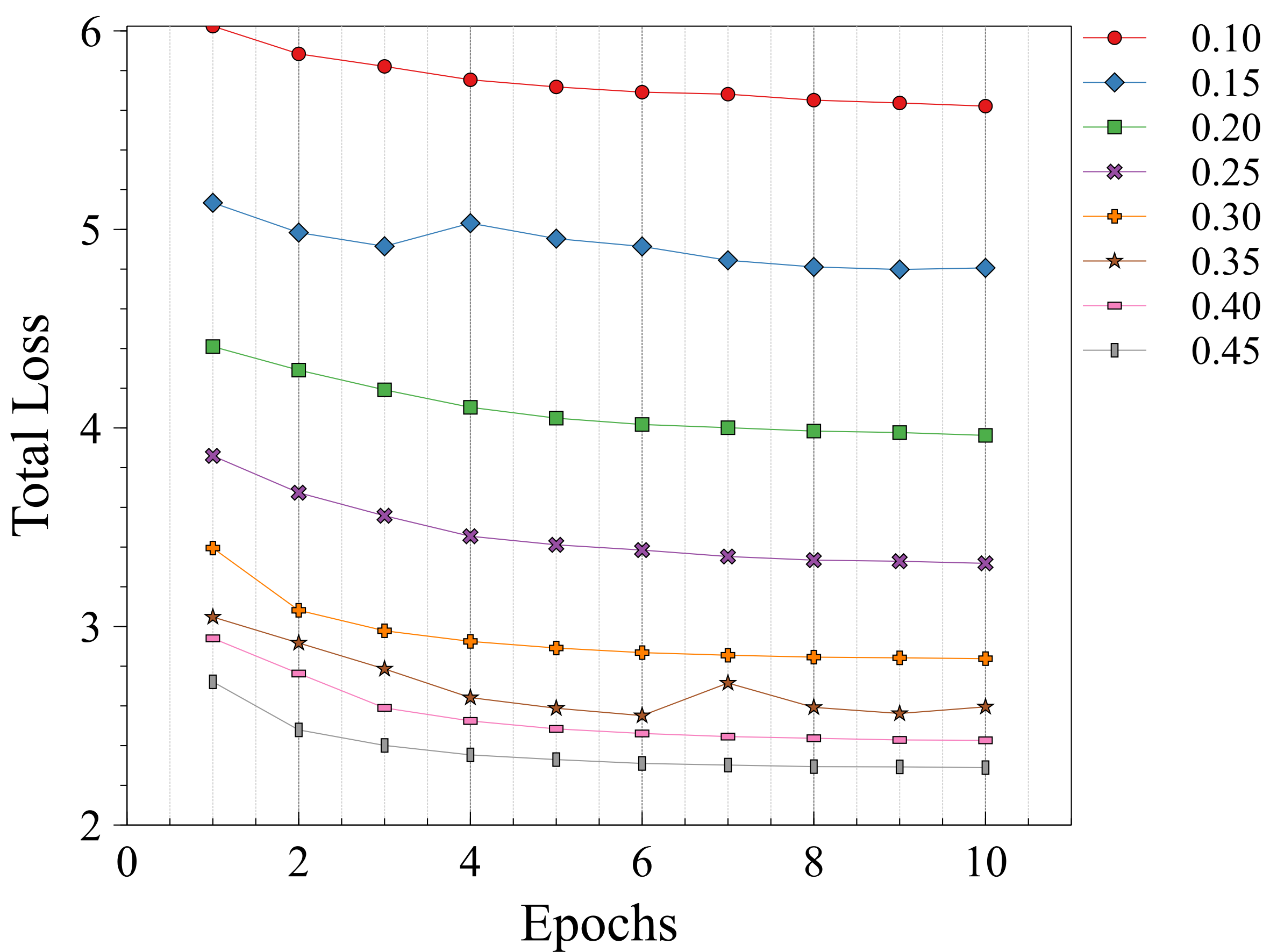}
    \caption{The total loss values during training for \textit{FlowReg-O} at different $\alpha$ values in the Charbonnier penalty function (Eqn. \ref{eqn: charbonnier}).}
    \label{fig:alpha_losses}
\end{figure}

\begin{figure}[!h]
    \centering
    \includegraphics[width=\textwidth]{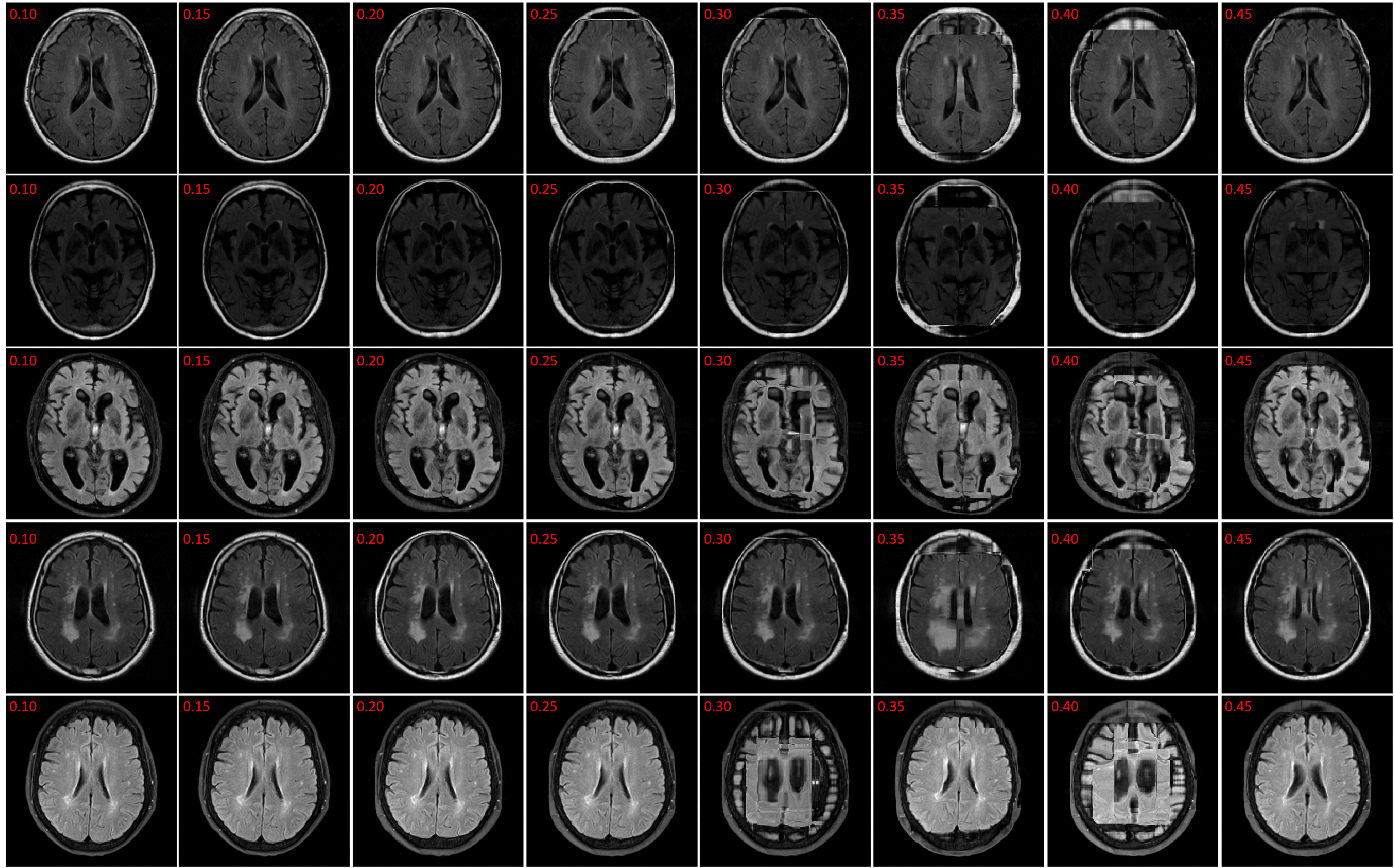}
    \caption{Single slices from five volumes registered using \textit{FlowReg-O} at various $\alpha$ values for the Charbonnier penalty.}
    \label{fig:slices_alpha}
\end{figure}

\begin{figure}[!h]
    \centering
    \includegraphics[width=\textwidth]{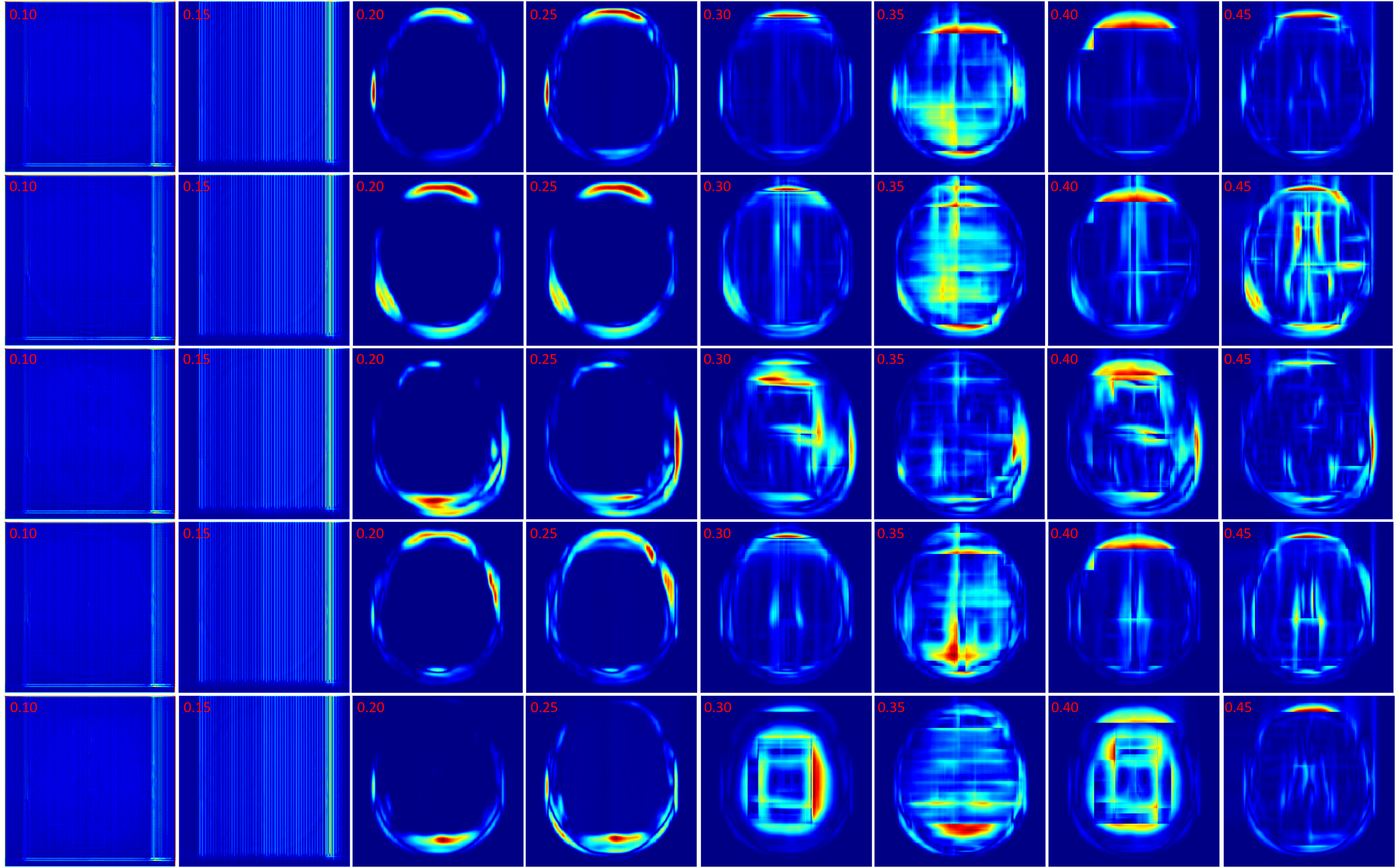}
    \caption{Flow magnitudes of deformation fields from \textit{FlowReg-O} at various $\alpha$ values. Images correspond to slices in Figure \ref{fig:slices_alpha}. Blue indicates areas of low flow vector magnitude and red indicates larger vector magnitude.}
    \label{fig:flow_mag_img}
\end{figure}

\clearpage

\section*{Appendix B}
\label{sec:appendixB}
\begin{figure}
    \centering
    \includegraphics[width=0.75\textwidth]{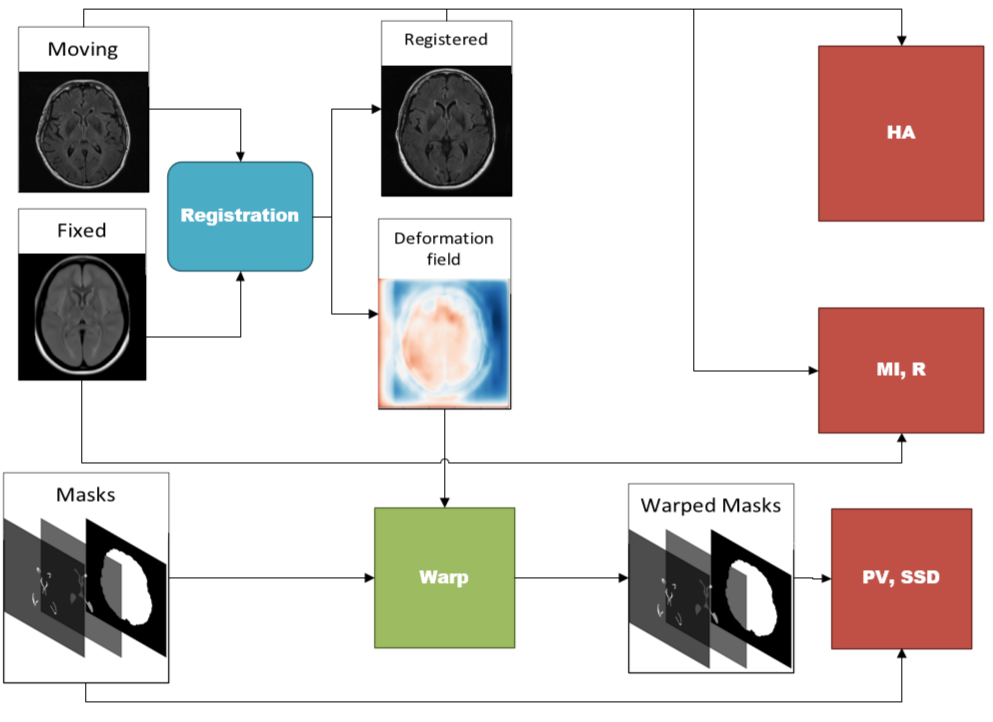}
    \caption{Registration metrics extraction process. HA = Head Angle, MI = Mutual Information, R = Correlation Coefficient, PV = Proportional Volume, SSD = Surface to Surface Distance.}
    \label{fig:process}
\end{figure}

Here we describe in detail the calculations of each registration metric.  The block diagram to compute the metrics is shown in Figure \ref{fig:process}. 
\subsection*{Structural Integrity: Proportional Volume --- PV}
If a registration scheme maintains the structural integrity of anatomical and pathological objects, the relative volume of objects should remain approximately the same after registration. Using binary masks of anatomical or pathological objects of interest, the proportional volume (PV) is proposed to measure the volume of a structure ($vol_s$) compared to the total brain volume ($vol_b$), as in:
\begin{equation}
    \label{eqn: PV}
    PV = \frac{vol_s}{vol_b}.
\end{equation}
The volume of objects in physical dimensions is found by multiplying the number of pixels by voxel resolution:
\begin{equation}
    \label{eqn: volume}
    vol = V_x \times V_y \times V_z \times n_p,
\end{equation}
where $vol$ is the volume in $mm^3$, $V_x$ and $V_y$ are the pixel width and height and $V_z$ is the slice-thickness. \\
\indent The difference between the PV before and after registration can be investigated to analyze how registration changes the proportion of each structure with respect to the brain. The difference in PV before and after registration can be measured by 
\begin{equation}
    \label{eqn: PV_change}
    \Delta PV = PV_{orig} - PV_{reg}.
\end{equation}
where $PV_{orig}$ and $PV_{reg}$ are the PV computed before and after registration.  In this work, two structures of interest are examined for $vol_s$: white matter lesions (WML) and ventricles since these objects are important for disease evaluation. Ideally the ratio of object volumes to the total brain volume would stay the same before and after registration 
\subsection*{Structural Integrity: Volume Ratio --- $\Delta V_s$}
In addition to the $\Delta PV$ which looks at proportional changes in volumes before and after registration, the volume ratio $\Delta V_s$ is also defined on a per object basis.  The volume ratio investigates the volumes of objects before and after registration, as in
\begin{equation}
    \label{eqn: volratio}
    \Delta V_s = \frac{vol_{orig}}{vol_{reg}},
\end{equation}
where $vol_{orig}$ and $vol_{reg}$ are the volumes of object $s$ before and after registration, respectively. The volume ratio is computed for three objects $s$: the brain, WML and ventricles.  This metric quantifies the overall change in volume before and after registration (the best ratio is a value equal to 1). 
\subsection*{Structural Integrity: Surface to Surface Distance --- SSD}
A third structural integrity measure, SSD, is proposed to measure the integrity \textit{between} objects in the brain before and after registration. In particular, the SSD measures how the overall shape of a structure changes after registration. To compute the SSD, binary masks of the brain $B(x,y,z)$, and structure $S(x,y,z)$ of interest are obtained, and an edge map is obtained to find $E_{B}(x,y,z)$ and $E_{S}(x,y,z)$, respectively.  For every non-zero pixel coordinate $(x, y, z)$ in the boundary of the structure, i.e. $E_S(x,y,z)=1$, the minimum Euclidean distance from the structure's boundary to the brain edge is found. This closest surface-to-surface distance between pixels in the objects' boundaries is averaged over the entire structure of interest to compute the average SSD 
\begin{equation}
    \label{eqn: SSD}
    SSD = \frac{1}{N}\left[\sum_{i=1}^{N}\mathrm{argmin}_{(x_s,y_s,z_s)\in S} \left(\sqrt{p + q + r}\right) \right],
\end{equation}
where $p=(x_s - x_b)^{2}$, $q=(y_s - y_b)^{2}$, $r=(z_s - z_b)^{2}$ are differences between points in the edge maps of $E_s$ and $E_b$, $(x_s,y_s,z_s)$ and $(x_b,y_b,z_b)$ are triplets of the co-ordinates in the binary edge maps for the structural objects of interest and the brain, respectively, and $N$ is the number of points in the structure's boundary.  The overall structural integrity of objects should be maintained with respect to the brain structure after registration, i.e. the distance between the objects and the brain shape should not be significantly altered. This metric can be used to investigate  the extent in which anatomical shapes are maintained or distorted by registration by examining the difference in SSD before and after registration by
\begin{equation}
    \label{eqn: SSD_change}
   \Delta SSD = \frac{SSD_{orig}-SSD_{reg}}{SSD_{orig}},
\end{equation}
where $SSD_{orig}$ and $SSD_{reg}$ is the SSD before and after registration, respectively.
\subsection*{Spatial Alignment: Head Angle --- HA}
Head Angle (HA) is an orientation metric that measures the extent to which the head is rotated with respect to the midsaggital plane. For properly registered data (especially for data that is being aligned to an atlas), the HA should be close to zero with the head being aligned along the midstagittal plane. To measure the HA, a combination of Principal Component Analysis (PCA) and angle sweep as described in \cite{rehman2018efficient, liu2001robust} is adopted to find the head orientation of registered and unregistered data. The MRI volume is first binarized using a combination of adaptive thresholding and opening and closing techniques to approximately segment the head. The coordinates of each non-zero pixel in this 2D mask are stored in two vectors (one for each coordinate) and the eigenvectors of these arrays are found through PCA.  The directions of eigenvectors specify the orientation of the major axes of the approximated ellipse for the head region in a slice with respect to the vertical saggital plane. Eigenvalues are the magnitude of the eigenvectors (or length of the axes). The largest eigenvalue dictates the direction of the longest axis which is approximately the head angle $\theta_1$.  For improved robustness to outliers and improve the precision of the estimated angles, a secondary refinement step is utilized to compute the refined HA $\theta_2$. Every second slice from the middle (axial) slice to the top of the head are used and the three smallest angles over all slices are taken as candidates for further refinement.The lowest angles are selected as they are the most representative of head orientation. Each selected slice is mirrored and rotated according to an angle sweep from $-2\times\theta_1<\theta_2<2\times\theta_1$ at $0.5\degree$ angle increments.  At every increment of the rotation, the cross-correlation between the mirrored rotating image and the original is calculated and a score is recorded. The angle at which the highest score is selected for the optimized value $\theta_2$. The final HA is obtained by summing the respective coarse and fine angle estimates, i.e. $\theta=\theta_1+\theta_2$. 
\subsection*{Spatial Alignment: Pixelwise Agreement --- PWA}
Physical alignment in 3D means that within a registered dataset, each subject should have high correspondence between slices, i.e. the same slice from each subject should account for the same anatomy across patients.  To measure this effect, a metric called Pixelwise Agreement (PWA) is proposed. It considers the same slice across all the registered volumes in a dataset and compares them to the same slice from an atlas template (the fixed volume) through the mean-squared error. The sum of the error is computed for each slice, to obtain a slice-wise estimate of the difference between the same slice in the atlas as compared to each of the same slices from the registered data: %
\begin{equation}
    \label{eqn: MSE}
    PWA(z) = \frac{1}{N_j}\frac{1}{N_{xy}} \sum_{j\in J} \sum_{(x,y)}(M_j(x,y,z)-F(x,y,z))^2
\end{equation}
where $z$ is the slice number for which PWA is computed, $M_j(x,y,z)$ is the moving test volume $j$ from a total of $N_j$ volumes from the dataset, $N_xy$ is the number of voxels in slice $z$ and $F(x,y,z)$ is the atlas. Thus, at each slice, for the entire dataset, the PWA compares every slice to the same slice of the atlas. Low PWA indicates high-degree of correspondence between all the slices from the registered dataset and that of the atlas and considers both spatial and intensity alignment.  If there is poor spatial alignment, there will be poor intensity alignment since different tissues will be overlapping during averaging.  The slice-based PWA may also be summed to get a total volume PWA, i.e. $\frac{1}{N_z}\sum_z PWA(z)$ where $N_z$ is the number of slices.
\subsection*{Spatial Alignment: Dice Similarity Coefficient --- DSC}
To further examine spatial alignment, manually delineated brain masks from the moving volumes $M(x,y,z)$ were warped with the calculated deformation and compared to the brain mask of the atlas template $F(x,y,z)$ through the Dice Similarity Coefficient (DSC):
\begin{equation}
    D S C=\frac{2|b_{M} \cap b_{F}|}{|b_{M}|+|b_{F}|},
\end{equation}
where $b_{M}$ is the registered, moving brain mask and $b_{F}$ is the brain mask of the atlas template. DSC will be higher when there is  a high-degree of overlap between the brain regions from the atlas and moving volume.  For visual inspection of overlap, all registered brain masks were averaged to summarize alignment accuracy as a heatmap.
\subsection*{Intensity Similarity: Mutual Information --- MI}
The first intensity-based metric used to investigate registration performance is the widely adopted Mutual Information (MI) metric that describes the statistical dependence between two random variables. If there is excellent alignment between the moving and fixed images, there will be tight clustering in the joint probability mass functions. The MI of two volumes $M(x,y,z)$ and $F(x,y,z)$ with PMFs of $p_M(i)$ and $p_F(i)$ is calculated as follows:
\begin{equation}
    \label{eqn: MI}
    I(M;F) = \sum_{f\in F}\sum_{m\in M} p_{(M,F)} (m,f) \log \left( \frac{p_{(M,F)}(m,f)}{p_{M}(m)  p_{F}(f)} \right)
\end{equation}
where $p_{(M,F)}(m,f)$ is the joint probability mass function of the intensities of the moving and fixed volumes, $p_{M}(m)$ is the marginal probability of the moving volume intensities, and $p_{F}(f)$ is the marginal probability for the fixed volume.
\subsection*{Intensity Similarity: Pearson Correlation Coefficient --- $r$}
The Pearson Correlation Coefficient, $r$, is used as the second intensity measure which quantifies the correlation between the intensities in the moving $M(x,y,z)$ and $F(x,y,z)$ volumes:
\begin{equation}
    \label{eqn: R}
    r(M,F) = \frac {\sum_{i=1}^{n} (M_i - \overline{M})(F_i -\overline{F})}{\sqrt{\sum_{i=1}^{n}(M_i - \overline{M})^2} \sqrt{\sum_{i=1}^{n}(F_i - \overline{F})^2}}
\end{equation}
where $N$ is the number of voxels in a volume, $M_i$ and $F_i$ are the pixels from the moving and fixed volumes, and $\overline{F}$ and $\overline{M}$ are the respective volume mean intensities. 
\subsection*{Intensity Similarity: Mean Intensity Difference --- MAID}
The last registration performance metric considered is the mean intensity difference, MAID, which measures the quality of a newly generated atlas $A(x,y,z)$ compared to the original atlas (fixed volume). To create the new atlas, the moving volumes $M(x,y,z)$ from a dataset are registered to the original atlas $F(x,y,z)$ and then the same slices are averaged across the registered dataset generating the atlas $A(x,y,z)$. The intensity histograms of the original $F(x,y,z)$ and newly generated atlases $A(x,y,z)$ are compared through the mean absolute error to get the MAID, as in
\begin{equation}
\label{eqn: Mean-Error}
MAID(A,F) = \frac{1}{N_i}\sum_i|p_A(i)-p_{F}(i)|
\end{equation}
where $p_F$ and $p_A$ are the probability distributions of the intensities for the fixed (original atlas) and moving volumes (new atlas) and $N_i$ is the maximum number of intensities.  Changes to the intensity distribution of registered images  could arise from poor slice alignment.  The higher the similarity between the new atlas and the original, the lower the error between the two.
\clearpage

\end{document}